\documentclass[11pt,a4paper]{article}
\usepackage[T1]{fontenc}
\usepackage[utf8]{inputenc}
\usepackage{lmodern,color}
\usepackage[margin=26mm,top=24mm,bottom=25mm]{geometry}
\usepackage{amsmath,amssymb,amsthm,mathtools}
\usepackage{microtype,booktabs,array}
\usepackage[round,authoryear]{natbib}
\usepackage[hidelinks]{hyperref}
\usepackage[nameinlink,noabbrev]{cleveref}
\usepackage{enumitem}
\setlist{nosep}
\allowdisplaybreaks[2]
\newtheorem{theorem}{Theorem}[section]
\newtheorem{proposition}[theorem]{Proposition}
\newtheorem{lemma}[theorem]{Lemma}
\newtheorem{corollary}[theorem]{Corollary}
\theoremstyle{definition}
\newtheorem{assumption}{Assumption}[section]
\newtheorem{example}{Example}[section]

\theoremstyle{remark}

\numberwithin{equation}{section}
\newcommand{\PP}{\mathbb P}
\newcommand{\EE}{\mathbb E}
\newcommand{\RR}{\mathbb R}
\newcommand{\NN}{\mathbb N}
\newcommand{\ZZ}{\mathbb Z}
\newcommand{\ind}{\mathbf 1}
\newcommand{\past}{E^{-}}
\newcommand{\TV}{\mathrm{TV}}
\newcommand{\Lip}{\mathrm{Lip}}
\newcommand{\clip}{\operatorname{clip}_{[0,1]}}
\newcommand{\diam}{\operatorname{diam}}

\newcommand{\cF}{\mathcal F}
\newcommand{\dl}{\Delta_L}

\newcommand{\norm}[1]{\left\lVert #1\right\rVert}
\newcommand{\abs}[1]{\left\lvert #1\right\rvert}
\title{Empirical concentration for chains with infinite memory}
\author{Pierre Alquier\\\small ESSEC Business School\footnote{This work was funded by CY Initiative, a programme supported the French government grant:
“Investissements d’avenir” \#France2030.}
\footnote{The authors acknowledge they used AI when preparing this manuscript. A preliminary version of the paper was submitted to AI, which suggested significant improvements for the results in the continuous case. The authors take full responsibility for the possible remaining mistakes.}
\and Paul Doukhan\\\small CY Cergy Paris Universit\'e$^{*\dagger}$}
\begin{document}
\maketitle
\begin{abstract}
Concentration inequalities for time series usually involve unknown quantities measuring dependence. This prevents their direct use as confidence bounds, or as criteria for learning algorithms. Recent works proved empirical concentration inequalities for stationary Markov chains, where the unknown quantities are estimated. We extend this approach beyond the Markov case, by considering  stationary chains with infinite memory. We first establish a bounded-differences inequality on bounded Polish metric spaces under a summable Wasserstein contraction condition. We then estimate its dependence coefficients from a single trajectory, using either transition counts when the state space is finite, or derivatives of kernel conditional-density estimates for a continuous state space. We give general error bounds and illustrate their dependence on the memory tail. Finally, a fixed multiplicative safety factor turns these estimators into empirical concentration bounds. While some regularity assumptions are required to make the bounds empirical, our procedure remains fully adaptive in the sense that it does not require the knowledge of the constants involved in these assumptions. The empirical confidence bounds derived from our results are asymptotically as tight as the theoretical ones.
\end{abstract}

\section{Introduction}\label{sec:intro}
Suppose that we observe a sequence of random variables and wish to estimate the expectation of a bounded function. Concentration inequalities tell us how far its empirical average can be from its expectation. For independent observations, Hoeffding's inequality gives a confidence interval that can be computed as soon as the range of the function is known~\citep{BLM2013}. This simple fact is also useful well beyond the estimation of a mean: it is one of the ingredients of generalization bounds and model selection procedures~\citep{mohri2012foundations,bach2024learning}, and upper-confidence-bound algorithms for bandits~\citep{LattimoreSzepesvari2020}.

For dependent observations, there is an additional difficulty. Concentration inequalities generally involve a quantity describing the dependence: for example, a sum of mixing coefficients, a contraction coefficient, a mixing time, or a spectral gap. Even when the inequality is explicit, this quantity need not be known. Such a bound is a valuable mathematical tool to control the probability of deviations from the mean in theory, but it cannot directly be evaluated on a sample. In particular, using such a bound to build a criterion to model selection, or confidence intervals for UCB-type algorithms, requires additional information about the data-generating process. This issue already appears in model selection for time series forecasting, as in~\citet{mohri10a,AlquierWintenberger2012}, among others.

A natural way to remove this difficulty is to estimate the dependence quantity as well. For exmaple, Paulin's version of Bernstein inequality for Markov chains depends on the pseudo-spectral gap~\citep{Paulin2015}. Estimators of the pseudo-spectral gap were recently developped, see~\cite{WolferKontorovich2024} for finite state-spaces, and~\cite{KaragulyanKasprzakAlquierDraft} for continuous, compact state-spaces. Combining these tools allows empirical deviation inequalities, and thus generalization bounds~\citep{KaragulyanAlquier2026} or confidence bounds in UCB-type algorithms~\citep{KaragulyanKasprzakAlquierDraft}. In both cases, the non-asymptotic nature of the initial concentration inequality is lost: however, the "semi-asymptotic" result obtained turns out to be enough for the application to generalization bounds and to bandits.

The objective of this paper is to extend this approach beyond the Markov case: we consider stochastic processes where the distribution of the next observation may depend on the entire past. We assume that the effect of changing the observation at lag $j$ is bounded by a coefficient $\delta_j(P)$, and that
\[
 \Delta(P):=1-\sum_{j\ge1}\delta_j(P)>0
\]
(we will provide a formal definition below, in Section~\ref{sec:setting}).
The number $\Delta(P)$ is a \emph{contraction margin}: a small margin allows a perturbation to have a large cumulative effect. Our first result is a McDiarmid type concentration inequality with a dependence penalty $1/\Delta(P)$. In the case of a discrete state space, our result recovers essentially an inequality from~\cite{ChazottesGalloTakahashi2023}. We then estimate the margin on a finite alphabet and on a compact, continuous state-space $[0,1]^d$.

To obtain a confidence interval with the correct sample-size scale, we do not have to estimate $\Delta(P)$ at a parametric rate. This is already pointed out by~\cite{KaragulyanAlquier2026}. We do not even need a known, shrinking upper bound on its estimation error. Suppose that $\widehat\Delta_n$ is accurate at every fixed tolerance, with a failure probability eventually smaller than any prescribed inverse power of $n$. Then, for a fixed $\kappa>0$,
\[
 \underline\Delta_n=\frac{\clip(\widehat\Delta_n)}{1+\kappa}
\]
is eventually a lower confidence bound on $\Delta(P)$, where $\clip(x):=\max(0,\min(x,1))$. The radius obtained by substituting $\underline\Delta_n$ in the concentration inequality is asymptotically $1+\kappa$ times the oracle radius. The price is an unknown, possibly large, sample size before the guarantee becomes valid, but no unknown constant enters the definition of the procedure.

\subsection{Related works}
There is a substantial literature on concentration without independence. Transportation and coupling approaches for contracting Markov chains were developed by \citet{Marton1996}; \citet{Samson2000} also treats uniformly mixing processes. \citet{Rio2000} proves Hoeffding-type inequalities for Lipschitz functions of dependent sequences. Bernstein inequalities under strong mixing are studied by \citet{MerlevedePeligradRio2009}. General accounts of weak dependence and its statistical consequences can be found in \citet{DedeckerEtAl2007}. The martingale approach is developed by \citet{KontorovichRamanan2008} and, in a general metric framework using Wasserstein matrices, by \citet{KontorovichRaginsky2017}. These works provide much of the probabilistic background for our argument.

For Markov chains, \citet{Paulin2015} gives bounded-differences and Bernstein inequalities, using mixing times and the pseudo-spectral gap. \citet{DedeckerFan2015,fan2022deviation} study separately Lipschitz functionals of contractive iterated random functions. The infinite-memory setting considered here is related to the causal models of \citet{DoukhanWintenberger2008}. Most directly, \citet{ChazottesGalloTakahashi2023} proves Gaussian concentration on countable alphabets under conditions on the oscillations or variations of the transition kernel. \citet{DoukhanFan2026} obtains deviation and moment inequalities for contractive infinite-memory processes using a random-function representation. Our concentration theorem uses a kernel-level Wasserstein condition and treats the stationary remote past explicitly. The point of the formulation here is its explicit dependence penalty, which we subsequently estimate.

The estimation of dependence from one trajectory is more recent. \citet{McDonaldShaliziSchervish2015} studies histogram estimation of $\beta$-mixing coefficients. \citet{KhaleghiLugosi2023} constructs consistent estimators of mixing coefficients and their sums. For Markov processes, see also \citet{GrunewalderKhaleghi2024,WolferAlquier2024}. A particularly useful line of work estimates mixing times, spectral gaps, and contraction coefficients together with data-dependent confidence intervals: \citet{HsuEtAl2019,WolferKontorovich2019,WolferKontorovich2024,Wolfer2024}. These results, together with the standard PAC-Bayesian toolbox~\citep{Alquier2024}, allowed \citet{KaragulyanAlquier2026,KaragulyanKasprzakAlquierDraft} to derive generalization bounds. They also motivated this work.

We also mention an alternative approach, based on regeneration times. \citet{Jerison2026} used this approach to prove a Dvoretzky–Kiefer–Wolfowitz inequality for regenerative Markov chains. A very nice feature of this work is that it does not require the estimation of any dependence coefficient. One drawback is that it leads to deviation inequalities whose first order terms are empirical, but second order terms are not. This is not a limitation for applications in machine learning, as shown by~\citet{Imon}.

For clarity, we mention that the word ``empirical'' is used in different senses in this literature. For example, empirical Bernstein inequalities replace an unknown variance by an empirical quantity; this does not necessarily eliminate unknown dependence coefficients when the data is not independent. The inequalities of \citet{MirzaeiEtAl2025} concern dependent Hilbert-valued observations and retain dependence information in their calibration. Here, \emph{empirical} means that the formula for the concentration bound requires no numerical knowledge of the process, apart from the known state space and the regularity of the statistic being evaluated.

For continuous observations, our estimator differentiates a kernel estimate of a conditional density. This is related to the ratio methods of \citet{DoukhanLang2009} and to the time-series kernel theory of \citet{WuHuangHuang2010}.

The rest of the paper follows the construction of the empirical bound. Section~\ref{sec:setting} introduces the setting. Section~\ref{sec:concentration} gives the concentration inequality and explains the statistical objective. Sections~\ref{sec:discrete} and~\ref{sec:continuous} estimate the dependence penalty in the discrete and continuous cases respectively. All proofs are in Section~\ref{sec:proofs}. We do not develop applications to PAC-Bayes or UCB here, as they would be essentially similar to~\cite{KaragulyanKasprzakAlquierDraft}.

\section{Setting and notation}\label{sec:setting}
We observe $(X_1,\ldots,X_n)$, a segment of a stationary process $(X_t)_{t\in\ZZ}$ with values in a Polish metric space $(E,\rho)$. We follow the conventions of~\citet{ChazottesGalloTakahashi2023} and will index the past by $\past=E^{\{-1,-2,\ldots\}}$: for $u\in\past$, $u=(u_{-1},u_{-2},\ldots)$. We fix a measurable probability kernel $u\mapsto P_u$ such that, almost surely,
\[
 \mathcal L(X_t\mid X_{t-1},X_{t-2},\ldots)
 =P_{(X_{t-1},X_{t-2},\ldots)}.
\]
For probability measures $\mu,\nu$ on $E$, let
\[
 W_{1,\rho}(\mu,\nu)=\inf_{\gamma\in\Gamma(\mu,\nu)}
 \int\rho(x,y)\,\gamma({\rm d}x,{\rm d}y),
\]
where $\Gamma(\mu,\nu)$ is the set of their couplings. On a countable space with the discrete (Hamming) metric, this is the total variation distance
$\norm{\mu-\nu}_{\TV}=\sup_A|\mu(A)-\nu(A)|$.

\begin{assumption}[Bounded state space]\label{ass:bounded}
The state-space is bounded,
$\Delta_E:=\diam(E,\rho)<\infty$.
\end{assumption}
\begin{assumption}[Wasserstein contraction]\label{ass:contract}
There are nonnegative coefficients $(\delta_j(P))_{j\ge1}$ such that, for all $u,v\in\past$,
\begin{equation}\label{eq:contract}
 W_{1,\rho}(P_u,P_v)\le\sum_{j\ge1}\delta_j(P)\rho(u_{-j},v_{-j})\text{ and }\Delta(P):=1-\sum_{j\ge1}\delta_j(P)>0.
\end{equation}
\end{assumption}
These two assumptions will be enough for our deviation inequality. However, when we turn to the estimation of $\Delta(P)$, additional regularity and positivity conditions will be required. As they will depend on whether the state-space if discrete or continuous, they will be discussed later, in the corresponding sections.

Also note that the coefficients in \eqref{eq:contract} need not be unique. The concentration theorem holds for any admissible sequence. For estimation, however, we must specify a target: Section~\ref{sec:discrete} uses minimal total-variation oscillations, while Section~\ref{sec:continuous} uses minimal density-Lipschitz coefficients and a fixed conversion to Wasserstein distance.

We often suppress the argument $P$ when it is clear from the context: $\Delta:=\Delta(P)$ and $\delta_j:=\delta_j(P)$.
For $(a,b)\in\mathbb{R}^2$ we write $a_+=\max(a,0)$, $a\vee b=\max(a,b)$, $a\wedge b = \min(a,b)$ and $\clip(a)=0 \vee (a\wedge 1)$. When our upper bounds are fractions, they are interpreted as $+\infty$ when the denominator is $0$.

\section{Concentration inequality and objective}\label{sec:concentration}
\begin{theorem}
\label{thm:concentration}
Under Assumptions~\ref{ass:bounded} and~\ref{ass:contract}, let $F:E^n\to\RR$ be separately Lipschitz with constants $c_1,\ldots,c_n\ge0$: replacing $x_j$ by $x_j'$ changes $F(x_1,\dots,x_n)$ by at most $c_j\rho(x_j,x'_j)$. Then, for every $s\in\RR$,
\begin{equation}\label{eq:mgf}
 \EE\exp \Bigl\{s\bigl[F(X_1,\ldots,X_n)-\EE F(X_1,\ldots,X_n)\bigr] \Bigr\}
 \le\exp\left\{\frac{s^2\Delta_E^2}{8\Delta(P)^2}
                    \sum_{j=1}^n c_j^2\right\}.
\end{equation}
Set $F_n=F (X_1, \ldots, X_n) $ for simplicity. Then for any $\delta\in(0,1)$ we obtain,
\begin{equation}\label{eq:oracle}
 \PP\left( |F_n-\EE F_n|\le
 \frac{\Delta_E}{\Delta(P)}
 \sqrt{\frac12\sum_{j=1}^n c_j^2\log\frac2\delta}\right)\ge1-\delta.
\end{equation}
\end{theorem}
For independent observations, we can take $\delta_j=0$ for every $j$ and recovers the usual bounded-differences inequality \citep{McDiarmid1989}, with exactly the same constants. On a discrete alphabet, the dependence on $\Delta(P)$ is the one in the oscillation-based result of \citet{ChazottesGalloTakahashi2023}. We remind that all the proofs are in Section~\ref{sec:proofs}.

We now exemplify a typical application of such a bounded-difference inequality to machine learning. This will allow to clarify the problems raised by the dependence of the bound in $\Delta(P)$. In learning problems, we need to control deviations of the empirical risk of a predictor $\theta$, $\widehat R_n(\theta)$, from its population risk $R(\theta)$:
\[
 \widehat R_n(\theta)=\frac1n\sum_{i=1}^n\ell_\theta(X_i),
 \qquad R(\theta)=\EE[\ell_\theta(X_0)]
\]
where $\ell_\theta(\cdot)$ is a suitable loss function. Assume $\ell_\theta$ takes values in $[0,1]$. In this case,
$
F_n = \widehat R_n(\theta) 
$
satisfies the assumptions of Theorem~\ref{thm:concentration} with $c_j=1/n$.
Thus \eqref{eq:oracle} gives
\begin{equation}\label{eq:riskradius}
 \PP\left( |  \widehat R_n(\theta) - R(\theta)|\le
 \frac{\Delta_E}{\Delta(P)}
 \sqrt{\frac{\log\frac2\delta}{2n}}\right)\ge1-\delta.
\end{equation}
 The numerator in $\Delta_E/\Delta(P)$ in~\eqref{eq:oracle} is usually known in practice, it depends on the type objects that are used as inputs in the learning algorithm. But it is not reasonable to assume that the denominator is known. Our objective is therefore to replace $\Delta(P)$ by an observable lower bound. A two-sided confidence interval for $\Delta(P)$ would be sufficient, but the following proposition shows we actually need a little less.

\begin{proposition}
\label{prop:safety}
Let $\widetilde\Delta_n\in[0,1]$ be an observable statistic. Suppose that, for every $\varepsilon>0$ and $q>0$, there is $N(P,\varepsilon,q)<\infty$ such that, for any $n\ge N(P,\varepsilon,q)$,
\begin{equation}\label{eq:fixedaccuracy}
 \PP_P(|\widetilde\Delta_n-\Delta(P)|>\varepsilon)\le n^{-q}.
\end{equation}
Fix $0<\kappa\le1$ and put
\begin{equation}\label{eq:safety}
 \underline\Delta_n=\frac{\widetilde\Delta_n}{1+\kappa}.
\end{equation}
For every $q>0$, there is $N(P,\kappa,q)<\infty$ such that, for $n\ge N(P,\kappa,q)$,
\begin{equation}\label{eq:margincoverage}
 \PP_P\left(\frac{\Delta(P)}{2(1+\kappa)}
             \le\underline\Delta_n\le\Delta(P)\right)\ge1-n^{-q}.
\end{equation}
\end{proposition}
For such $n$, a union bound on~\eqref{eq:oracle} and~\eqref{eq:margincoverage} gives:
\begin{equation}\label{eq:oracle:empirical}
 \PP\left( |F_n-\EE F_n|\le
 \frac{\Delta_E}{\underline\Delta_n}
 \sqrt{\frac12\sum_{j=1}^n c_j^2\log\frac2\delta}\right)\ge1-\delta-n^{-q}.
\end{equation}
As mentioned at the end of Section~\ref{sec:setting}, we allow $\underline\Delta_n=0$ by taking the deviation radius as infinite in this case. Formally, for $c=(c_1,\dots,c_n) $ and $x\in[0,1]$ we put $\mathcal{R}_{n,\delta}(c,\Delta_E,x) = +\infty$ if $x=0$, while for  $x>0$:
$$
\mathcal{R}_{n,\delta}(c,\Delta_E,x) := \frac{\Delta_E}{x}
 \sqrt{\frac12\sum_{j=1}^n c_j^2\log\frac2\delta}
$$
so that we can write~\eqref{eq:oracle:empirical} rigourously:
$ \PP\left( |F_n-\EE F_n|\le \mathcal{R}_{n,\delta}(c,\Delta_E,\underline\Delta_n) \right)\ge1-\delta-n^{-q} $.
If $\widetilde\Delta_n\to\Delta(P)$ almost surely, the ratio of the empirical deviation radius to the theoretical one tends almost surely to $1+\kappa$:
$$
\frac{\mathcal{R}_{n,\delta}(c,\Delta_E,\underline\Delta_n)}
{ \mathcal{R}_{n,\delta}(c,\Delta_E,\Delta(P))}
\xrightarrow[n\rightarrow\infty ]{a.s.} 1+\kappa,
$$
making the empirical bound asymptotically almost as tight as the theoretical one.
We remind that~\eqref{eq:oracle:empirical} is an eventually valid empirical inequality, not a guarantee valid uniformly over all processes at every $n$. Before the unknown threshold $N(P,\kappa,q)$, a finite radius is not certified by the theorem. Of course, it would be amazing to remove this limitation (and also more difficult)! However, our previous work on Markov chains~\cite{KaragulyanAlquier2026} shows that this is enough to prove regret bounds for UCB or generalization bounds for classification.

From now, we will exemplify the coefficients $\delta_j(P)$, and discuss their estimation, first in the discrete case (Section~\ref{sec:discrete}) and later in the continuous case (Section~\ref{sec:continuous}).

\section{The discrete case}\label{sec:discrete}
\subsection{Assumptions and examples}
Let $E$ be a finite alphabet of known size $M\ge2$, equipped with the Hamming metric $\rho(x,y)=\ind_{\{x\ne y\}}$. Write $p(x\mid u):=P_u(\{x\})$. We use the canonical coefficients
\begin{equation}\label{eq:canonicaldiscrete}
 \delta_j(P)=\sup_{\substack{u,v\in\past\\u_{-l}=v_{-l}\ (l\ne j)}}
     \frac12\sum_{x\in E}|p(x\mid u)-p(x\mid v)|\text{ and  } T_K(P):=\sum_{j>K}\delta_j(P).
\end{equation}
We assume that $p$ is continuous in the product topology and $\sum_j\delta_j(P)<1$. Telescoping one coordinate at a time gives \eqref{eq:contract}. Continuity is important here: a condition on single-coordinate changes alone does not rule out dependence on a tail event. Also, note that as for $\Delta(P)$ and $\delta_j(P)$, we will sometimes drop the dependence in $P$ from $T_K(P)$: $T_K=T_K(P)$.

While Theorem~\ref{thm:concentration} holds without any further assumption, the estimation of $\Delta(P)$ will require the following.
\begin{assumption}\label{ass:nonnull}
There exists a $\pi_*>0$ (unknown) such that
$\inf_{x\in E,\,u\in\past}p(x\mid u)\ge\pi_*$.
\end{assumption}
This ensures that every finite word then has positive probability, making the continuous kernel, and the coefficients in \eqref{eq:canonicaldiscrete} identifiable. When writing that $\pi_*$ is unknown, we mean that the value of $\pi_*$ will only appear in the analysis, it is not used to compute our estimator of $\Delta(P)$.

We finally provide a few examples satisfying the assumptions.
\begin{example}[Markov chains]
For $p(x\mid u)=Q(u_{-1},x)$, all coefficients but $\delta_1$ vanish and
$\delta_1=\sup_{a,b}\norm{Q(a,\cdot)-Q(b,\cdot)}_{\TV}$.
If $Q(a,x)\ge\pi_*$ for all $a,x$, that is, Assumption~\ref{ass:nonnull} holds, then $\delta_1\le1-M\pi_*<1$: the chain is then uniformly ergodic, and satisfies Assumption~\ref{ass:contract}. We can also consider higher-order Markov chains. However, for a Markov-chain of order $q>1$, even under Assumption~\ref{ass:nonnull}, it is possible that $\delta_1+\dots+\delta_q \geq 1$. In this case, we really need Assumption~\ref{ass:contract} for Theorem~\ref{thm:concentration}.
\end{example}
\begin{example}
\label{ex:copy}
Let $(a_j)_{j\in\mathbb{N}}$ be nonnegative with $A=\sum_j a_j<1$, and let $\nu$ be a strictly positive probability on $E$. Set
\[
 p(x\mid u)=(1-A)\nu(x)+\sum_{j\ge1}a_j\ind_{\{u_{-j}=x\}}.
\]
At each time, the chain either draws a fresh observation from $\nu$, or copies one of its past observations. Here $\delta_j(P)=a_j$ and one may take $\pi_*=(1-A)\min_x\nu(x)$. This gives examples with finite, geometric, or polynomial memory tails.
\end{example}
\begin{example}
\label{ex:logistic}
For $E=\{0,1\}$, consider
\[
 p(1\mid u)=\sigma\left(b+\sum_{j\ge1}a_j u_{-j}\right),
 \qquad \sigma(z)=(1+e^{-z})^{-1}.
\]
If $\sum_j|a_j|<\infty$, non-nullness holds, and $\delta_j(P)\le |a_j|/4$. Hence $\sum_j|a_j|<4$ is a sufficient contraction condition. More generally, if $X_t\mid u$ has a binomial distribution $\mathcal{B}in(N,\sigma(b+\sum_j a_j u_{-j}))$ on $\{0,\ldots,N\}$, then $\delta_j(P)\le N^2|a_j|/4$.
\end{example}

\subsection{Estimator and a general error bound}
Fix $1\le K\le n/2$ and write $U_i=(X_{i-1},\ldots,X_{i-K})$. For $u\in E^K$, let
\[
 C_n(u)=\sum_{i=K+1}^n\ind_{\{U_i=u\}} \text{ and }
 \widehat p_K(x\mid u)=
 \frac{\sum_{i=K+1}^n\ind_{\{U_i=u,\,X_i=x\}}}{C_n(u)}.
\]
So $\widehat p_K(x\mid u)$ is defined only when $ C_n(u)>0$. Formally, we can extend the definition arbitrarily when $C_n(u)=0$, for example by taking $\widehat p_K(\cdot\mid u)$ to be the uniform probability on $E$ in that case. This is irrelevant in practice, however, no guarantees on the estimation of $\Delta(P)$ will be available in that situation.

Replace $p$ by $\widehat p_K$ to estimate the oscillations:
\begin{align}
 \widehat\delta_{j,K}
 &=\max_{\substack{u,v\in E^K\\u_{-l}=v_{-l}\ (l\ne j)}}
      \frac12\sum_{x\in E}|\widehat p_K(x\mid u)-\widehat p_K(x\mid v)|,
      \quad 1\le j\le K,\label{eq:discreteestimator}\\
 \widetilde\Delta_{n,K}
 &=\clip\left(1-\sum_{j=1}^K\widehat\delta_{j,K}\right).
 \label{eq:discretemargin}
\end{align}

\begin{theorem}
\label{thm:discretebound}
Put
\[
 H_{n,K}(\eta):=\log\frac{2^{M+1}M^K(n-K)}{\eta} \text{, and }
 r_{n,K}(\eta):=\max_{u\in E^K}
       \sqrt{\frac{H_{n,K}(\eta)}{2C_n(u)}}.
\]
For any $\eta\in(0,1)$,
\begin{equation}\label{eq:discreteobserved}
 \PP\left(|\widetilde\Delta_{n,K}-\Delta(P)|
      \le 2Kr_{n,K}(\eta)+(2K+1)T_K\right)\ge1-\eta.
\end{equation}
Under Assumption~\ref{ass:nonnull}, with probability at least $1-\eta-\mathcal R_{n,K}(P)$,
\begin{equation}\label{eq:discretedeterministic}
 |\widetilde\Delta_{n,K}-\Delta(P)|
 \le 2K\sqrt{\frac{H_{n,K}(\eta)}{(n-K)\pi_*^K}}+(2K+1)T_K,
\end{equation}
where
\begin{equation}\label{eq:countfailure}
 \mathcal R_{n,K}(P)=KM^K
       \exp\left\{-\frac{\lfloor(n-K)/K\rfloor\pi_*^K}{8}\right\}.
\end{equation}
\end{theorem}
There are two sources of error in \eqref{eq:discreteobserved}. The first is statistical and depends on how often each context is observed: this is a variance term. The second is the cost of replacing an infinite past by $K$ lags and then summing $K$ estimated influences (bias term). The calibration of $K$ is discussed in the next subsection.

\subsection{Consistency, rates, and empirical concentration}
\begin{corollary}
\label{cor:discrete}
Under Assumption~\ref{ass:nonnull}, assume $KT_K\to0$ when $K\to\infty$. Let $K_n\to\infty$ be a sequence of integers with $K_n=o(\log n)$ and $K_n\le n/2$. Then $\widetilde\Delta_{n,K_n}\to\Delta(P)$ almost surely and satisfies \eqref{eq:fixedaccuracy}.
\end{corollary}
In particular,
\[
 \underline\Delta_n=\frac{\widetilde\Delta_{n,K_n}}{1+\kappa}
\]
gives the empirical inequality of Proposition~\ref{prop:safety}. Its definition requires neither $\pi_*$ nor a bound on $T_K$.
The condition $KT_K\to0$ is qualitative. For example, $\delta_j(P)\le Cj^{-r}$ with any $r>2$ is sufficient, but neither $C$ nor $r$ has to be known.

For a concrete parameter-free choice, take
\begin{equation}\label{eq:discretetuning}
 K_n=\max\left\{1,\min\left(\lfloor n/2\rfloor,
                 \lfloor\sqrt{\log(n+e)}\rfloor\right)\right\},\qquad n\ge2.
\end{equation}
For each fixed process and $q>0$, \eqref{eq:discretedeterministic} gives, with probability at least $1-n^{-q}$ for all sufficiently large $n$,
\begin{equation}\label{eq:discreterate}
 |\widetilde\Delta_{n,K_n}-\Delta(P)|
 \le C(P,q)\left\{
 \frac{\log n}{\sqrt n}
       e^{\frac12\log(1/\pi_*)\sqrt{\log n}}
       +K_nT_{K_n}\right\}.
\end{equation}
Using this, we deduce the following rates hold for our estimator estimator; constants in the rates may depend on the $P$.
\begin{center}
\begin{tabular}{@{}p{.48\linewidth}p{.45\linewidth}@{}}
\toprule
Memory assumption & Rate for $\widetilde\Delta_{n,K_n}$\\
\midrule
$T_K=0$ for $K\ge K_0$, with $K_0$ unknown & $n^{-1/2+o(1)}$\\[3pt]
$\delta_j(P)\le Ce^{-bj}$, $b>0$ & $(\log n)^{1/2}e^{-b\sqrt{\log n}}$\\[3pt]
$\delta_j(P)\le Cj^{-r}$, $r>2$ & $(\log n)^{-(r-2)/2}$\\
\bottomrule
\end{tabular}
\end{center}
Here $n^{-1/2+o(1)}$ means a bound of order $n^{-1/2+\epsilon}$ for every fixed $\epsilon>0$, beyond a process-dependent threshold. Super-exponential tails $\delta_j(P)\le Ce^{-bj^\gamma}$ with $\gamma>2$ would also give this nearly parametric rate with \eqref{eq:discretetuning}. Finally, note that we do not claim optimality of these parameter-free rates. To derive the minimax rate for the estimation of $\Delta(P)$ seems to be a difficult open problem. It is certainly of interest, however, it is not necessary to get empirical concentration as in~\eqref{eq:oracle:empirical}. What matters is to have a parameter-free $K_n$ in~\eqref{eq:discretetuning}.

\section{The continuous case}\label{sec:continuous}
\subsection{Assumptions and examples}
We now consider $E=[0,1]^d$ equipped with the Euclidean metric. This time, we assume that $P_u$ has a density $p(\cdot \mid u)$ with respect to the Lebesgue measure: $P_u({\rm d} x)=p(x\mid u)\,{\rm d}x$. We define:
\begin{equation}\label{eq:continuousL}
 L_j=L_j(P):=\sup_{x\in E,\,u\in\past}\norm{\nabla_{u_{-j}}\,p(x\mid u)}_2,
 \quad L=L(P):=\sum_{j\ge1}L_j \text{ and }  T_K^L=T_K^L(P):=\sum_{j>K}L_j.
\end{equation}
With continuity in the infinite past, this gives
\begin{equation}\label{eq:densitylip}
 |p(x\mid u)-p(x\mid v)|\le\sum_{j\ge1}L_j\norm{u_{-j}-v_{-j}}_2.
\end{equation}
We can relate the coefficients $L_j(P)$ to Assumption~\ref{ass:contract} thanks to the following proposition.
\begin{proposition}
 \label{prop:eq:densityW}
    For any $(u,v)\in (E^{-})^2$ we have
   \begin{equation}\label{eq:densityW}
 W_{1,2}(P_u,P_v)\le\frac1\pi\norm{p(\cdot\mid u)-p(\cdot\mid v)}_\infty.
\end{equation} 
\end{proposition}
Thus we can take $\delta_j(P)=L_j/\pi$ in \eqref{eq:contract}.
Of course, it is possible that Assumption~\ref{ass:contract} still holds for smaller values of $\delta_j(P)$. However, for the purpose of empirical concentration, we do not necessarily need the tightest coefficients. Thus, we will below study the estimation of $L_j$ (rather than the estimation of the tightest possible value for $\delta_j(P)$), and of
\begin{equation}\label{eq:continuousmargin}
 \dl(P):=1-\frac{L}{\pi}>0.
\end{equation}

The proof of Proposition~\ref{prop:eq:densityW} uses Neumann--Poincar\'e inequality on the cube. The proposition can actually be extended to any domain that satisfies a Neumann--Poincar\'e inequality with constant $C_E>0$ instead of $1/\pi$. For example, when $E$ is convex and bounded, one may use $C_E\le\Delta_E/\pi$~\citep{PayneWeinberger1960}. However, the density estimators we use below are tailored for the unit square.

In order to estimate $\dl(P)$, we require the following additional assumption. The condition $  p(x\mid u)\ge\alpha$ is quite similar to the discrete setting, but our estimator based on kernel will also require $p(x\mid u)$ to be smooth enough.
\begin{assumption}
\label{ass:continuous}
First, $L<\pi$.
The kernel $p$ is continuous on $E\times\past$. Its restrictions to finitely many scalar coordinates are twice continuously differentiable, with derivatives extending continuously to the boundary. All first and second scalar coordinate derivatives are bounded in absolute value by one finite (unknown) constant $H$, uniformly over the coordinates and over $(x,u)$. For each $j$, the map $(x,u)\mapsto\nabla_{u_{-j}}p(x\mid u)$ is continuous in the product topology. There exists an (unknown) $\alpha>0$ such that
\[
 p(x\mid u)\ge\alpha\quad\text{for all }(x,u)\in E\times\past.
\]
\end{assumption}

Estimation of the Lipchitz constants will be done by estimating the supremum of the differentials. This leads to the introduction of one more tail quantity: jor $j\le K$, define
\[
 \omega_{j,K}:=\sup_{\substack{x\in E,\,u,v\in\past\\
                     u_{-l}=v_{-l}\ (1\le l\le K)}}
 \norm{\nabla_{u_{-j}}\,p(x\mid u)-\nabla_{v_{-j}}\,p(x\mid v)}_2\text{, and } \Omega_K:=\sum_{j=1}^K\omega_{j,K}.
\]
It measures how the derivatives, rather than just the density, depend on the remote past. Set
\begin{equation}\label{eq:continuousbias}
 L_0=\sup_{x,u}\norm{\nabla_x\,p(x\mid u)}_2\text{, and } B_K=\Omega_K+\left(1+ \frac{2\sqrt d\,K(L_0+L)}{\alpha} \right)T_K^L.
\end{equation}
Our general bound keeps $B_K$ explicit. Product continuity of the first derivatives and summability of $(L_j)$ imply $\Omega_K\to0$; hence $KT_K^L\to0$ is sufficient for $B_K\to0$. Quantitative rates still require control of $\Omega_K$. For the smooth-index examples below, the stronger relation $\Omega_K\le C T_K^L$ follows from their structure.

\begin{example}
\label{ex:normal}
On $[0,1]$, consider the conditional density
\[
 q(x;m,s)=\frac{\exp\{-(x-m)^2/(2s^2)\}}
                 {\int_0^1\exp\{-(y-m)^2/(2s^2)\}\,{\rm d}y}\text{, and } p(x\mid u)=q\left(x;b+\sum_{j\ge1}a_j u_{-j},s\right),
\]
where $s>0$ and $A=\sum_j|a_j|<\infty$. This is a Gaussian location model conditioned to remain in the state space. The location parameter ranges over a compact interval $I$. If
$C_1=\sup_{x\in[0,1],m\in I}|\partial_m q(x;m,s)|$, then $L_j=C_1|a_j|$; thus $C_1A<\pi$ is sufficient for \eqref{eq:continuousmargin}. Positivity and all the smoothness conditions hold, and $\Omega_K\le C T_K^L$. Finite, geometric, and polynomial coefficient tails are all possible.
\end{example}
\begin{example}
\label{ex:nonlinear}
We generalize the previous example with a non-linear functional form for the mean. That is,
replace the location in Example~\ref{ex:normal} by
$m(u)=b+\sum_j a_j\psi(u_{-j})$, where $\psi\in C^2([0,1])$ and $0<c_\psi\le\psi'\le C_\psi$. For example, $\psi(x)=\tanh x$ works. The coefficients $L_j$ are bounded above and below by fixed positive multiples of $|a_j|$, and $\Omega_K\le C\sum_{j>K}|a_j|$. The assumptions hold when thid sum is small enough.
\end{example}
\begin{example}
\label{ex:scale}
A heteroscedastic version is obtained by taking
\[
 m(u)=b+\sum_j a_j u_{-j},\quad
 s(u)=s_0+\sum_j b_j u_{-j}\text{, and } p(x\mid u)=q(x;m(u),s(u)),
\]
with summable $(|a_j|+|b_j|)$ and $0<s_{\min}\le s(u)\le s_{\max}<\infty$. On the resulting compact parameter set,
$L_j\le C(|a_j|+|b_j|)$ and $\Omega_K\le C\sum_{j>K}(|a_j|+|b_j|)$. The two Gaussian score directions are linearly independent, which also gives a lower comparison with $|a_j|+|b_j|$. Thus $\Omega_K\le C'T_K^L$. A sufficiently small sum of these coefficients ensures contraction.
\end{example}
\begin{example}
\label{ex:mixture}
Let $q_0,q_1$ be distinct, strictly positive $C^2$ densities on $[0,1]$ and set
\[
 p(x\mid u)=[1-w(u)]q_0(x)+w(u)q_1(x)\text{, and }
 w(u)=\sigma\left(b+\sum_j a_j u_{-j}\right).
\]
This describes a mixture whose weight changes depending on the context (the past). Again $L_j=C_1|a_j|$ for a fixed $C_1>0$, and $\Omega_K\le C T_K^L$. The density $p$ is uniformly positive without knowing the minimum of either component density.
\end{example}
A stationary version exists for these continuous kernels on a compact state space, and the contraction condition gives uniqueness.

\subsection{Estimator}
Fix $K\le n/2$ and $0<h\le1/4$. Let $k$ be a fixed, nonnegative $C^2$ density supported on $[-1,1]$, for example $k(t)=c(1-t^2)^3_+$ with the normalizing constant $c$. For a vector $z\in\RR^r$, we define as usual the product kernel $\mathcal K_r(z)=\prod_{a=1}^r k(z_a)$. Write $U_i=(X_{i-1},\ldots,X_{i-K})\in E^K$. Define
\begin{align}
\label{eq:kernel:estim:1}
 \widehat f_{K,h}(x,u)&=
 \frac1{(n-K)h^{d(K+1)}}\sum_{i=K+1}^n
 \mathcal K_d\left(\frac{X_i-x}{h}\right)
 \mathcal K_{dK}\left(\frac{U_i-u}{h}\right),\\
 \widehat g_{K,h}(u)&=
 \frac1{(n-K)h^{dK}}\sum_{i=K+1}^n
 \mathcal K_{dK}\left(\frac{U_i-u}{h}\right)\text{, and }
 \widehat g^\tau_{K,h}=\widehat g_{K,h}\vee n^{-1}.
 \label{eq:kernel:estim:2}
\end{align}
The quotient rule suggests the estimator
\begin{equation}\label{eq:derivativeestimator}
 \widehat D_{j,K,h}(x,u)=
 \frac{\nabla_{u_{-j}}\widehat f_{K,h}(x,u)}{\widehat g^\tau_{K,h}(u)}
 -\frac{\widehat f_{K,h}(x,u)\nabla_{u_{-j}}\widehat g_{K,h}(u)}
             {[\widehat g^\tau_{K,h}(u)]^2}.
\end{equation}
The deterministic trimming level $n^{-1}$ avoids using the explicit lower bound on the density, $\alpha$, which is not assumed to be known. We evaluate the estimator on the interior cube $\mathcal I_{K,h}=[h,1-h]^{d(K+1)}$ and put
\begin{equation}\label{eq:continuousestimator}
 \widehat L_{j,K,h}=\sup_{(x,u)\in\mathcal I_{K,h}}
                      \norm{\widehat D_{j,K,h}(x,u)}_2
                      ,\text{ and }
 \widetilde\Delta^L_{n,K,h}
 =\clip\left(1-\frac1\pi\sum_{j=1}^K\widehat L_{j,K,h}\right).
\end{equation}
These are finite-dimensional optimizations of explicit functions. They define observable statistics, although we do not claim the global optimization is easy! The boundary restriction shrinks with $h$; its approximation error is included in the theorem.

\subsection{General bound and parameter-free choices}
\begin{theorem}
\label{thm:continuousbound}
Under Assumption~\ref{ass:continuous}, there is a constant $C_P\ge2$, depending only on  the process through $P$, $d$,  on the kernel $k(\cdot)$, and such that \eqref{eq:continuousbound} holds for $n\ge2K+2$, $0<h\le1/4$, and $0<\eta<1$. Put
\begin{align}
\nonumber
 V_{n,K,h}(\eta) & =(d(K+1)+1)^2\log\frac{e(n+1)}{\eta h} \text{, and }
 \\
 e_{n,K,h}(P,\eta) & =C_P^{K+1}\left[
 h+h^{-(d(K+1)+2)}\sqrt{\frac{V_{n,K,h}(\eta)}{n}}\right].
 \label{eq:continuouserror}
\end{align}
Then
\begin{equation}\label{eq:continuousbound}
 \PP\left(|\widetilde\Delta^L_{n,K,h}-\dl(P)|
          \le e_{n,K,h}(P,\eta)+\frac{B_K}{\pi}\right)\ge1-\eta.
\end{equation}
\end{theorem}
The two terms into brackets are the bias (due to smoothing) and the variance (statistical error). The bandwidth exponent in \eqref{eq:continuouserror} is conservative. It comes from applying Theorem~\ref{thm:concentration} to Lipschitz kernel sums, without a variance-sensitive kernel inequality or an additional mixing assumption. For fixed $K$, balancing the two terms gives the rate $(\log n/n)^{1/(2d(K+1)+6)}$. We do not claim any optimality on this rate.
\begin{corollary}
\label{cor:continuous}
Assume $KT_K^L\to0$. For $n\ge4$, choose
\begin{align}
 K_n&=\max\left\{1,\min\left(\lfloor(n-2)/2\rfloor,
                         \lfloor[\log(n+e)]^{1/3}\rfloor\right)\right\},
 \label{eq:continuoustuningK}\\
 h_n&=\min\left\{\frac14,(n+e)^{-1/[4(d(K_n+1)+3)]}\right\}.
 \label{eq:continuoustuningh}
\end{align}
Then $\widetilde\Delta^L_{n,K_n,h_n}\to\dl(P)$ almost surely and satisfies \eqref{eq:fixedaccuracy} with $\dl(P)$ instead of $\Delta(P)$.
\end{corollary}
 Hence
\[
 \underline\Delta_n^L=\frac{\widetilde\Delta^L_{n,K_n,h_n}}{1+\kappa}
\]
gives the empirical concentration inequality of Proposition~\ref{prop:safety}. No value of $\alpha$, $H$, $L$, or a memory-tail constant is used in \eqref{eq:continuoustuningK}--\eqref{eq:continuoustuningh}.

For every fixed $q>0$, the smoothing and statistical error at level $\eta=n^{-q}$ is eventually bounded by
$\exp\{-c_d(\log n)^{2/3}\}$ for some $c_d>0$ depending only on $d$. When $\Omega_K\le C T_K^L$, the following examples of \eqref{eq:continuousbound} result:
\begin{center}
\begin{tabular}{@{}p{.48\linewidth}p{.45\linewidth}@{}}
\toprule
Memory assumption & Rate for $\widetilde\Delta^L_{n,K_n,h_n}$\\
\midrule
$T_K^L=0$ for $K\ge K_0$, $K_0$ unknown & $\exp\{-c_d(\log n)^{2/3}\}$\\[3pt]
$L_j\le Ce^{-bj}$, $b>0$ & $(\log n)^{1/3}e^{-b(\log n)^{1/3}}$\\[3pt]
$L_j\le Cj^{-r}$, $r>2$ & $(\log n)^{-(r-2)/3}$\\
\bottomrule
\end{tabular}
\end{center}

\section{Conclusion}

We extended the approach developed by~\cite{KaragulyanKasprzakAlquierDraft} for Markov chains, to derive empirical concentration bounds for chains with infinite memory. While the rate of convergence can be slow, the empirical confidence interval remain asymptotically in $1/\sqrt{n}$. We thus have a proof of concept that empirical concentration bounds are possible for a large class of time series. While the optimality of the rates remains of course an interesting open question, we believe a crucial point is to go beyond the bounded case. To be frank, 
in the unbounded case, all our tools collapse. First, we can no longer have a uniform estimation of the condition densities, making it difficult to estimate the Lipschitz constants $L_j$. Moreover, even assuming the $X_i$'s remain sub-Gaussian, their sub-Gaussian norm $\|X_i\|_{\psi_2}$ will appear in the concentration inequality~\citep{kontorovich2014concentration}. Thus if we want to keep the bounds empirical, this norm should be also estimated. While this is possible, this cannot be done with rates of convergence in general~\citep{mies2026empirical}. If feasible at all, the unbounded case will probably require a different approach.

\section*{Acknowledgements}

This work was funded by CY Initiative, a programme supported the French government grant:
“Investissements d’avenir” \#France2030 (Project SafeTime). The second author also wishes to thank the Chair Actions \url{https://sites.google.com/view/chaireactions/} and  Gachon University in Seoul for their support.
Both authors thank Geoffrey Wolfer (Tokyo University of Agriculture and Technology), who introduced the authors to the paper~\cite{ChazottesGalloTakahashi2023}.

\section{Proofs}\label{sec:proofs}

These additional notations that will be used to make some formulas less cumbersome:
\begin{equation}
m  = n-K \text{ and }
D  = d(K+1).
\label{eq:notcumbersome}
\end{equation}

\subsection{Proofs of Section~\ref{sec:concentration}}

\begin{proof}[Proof of Theorem~\ref{thm:concentration}]
Put $a=\sum_{j\ge1}\delta_j=1-\Delta<1$. Define the renewal sequence
\begin{equation}\label{eq:renewal}
 r_0=1,\qquad r_\ell=\sum_{j=1}^\ell\delta_jr_{\ell-j}\quad(\ell \ge1).
\end{equation}
Extending $(\delta_j)_{j\ge1}$ to a sequence on $\NN_0=\{0,1,2,\dots\}$ by putting $\delta_0=0$, we have $r=\sum_{k\ge0}\delta^{*k}$. Tonelli's theorem therefore gives
\begin{equation}\label{eq:renewalmass}
 \sum_{\ell\ge0}r_\ell=\sum_{k\ge0}a^k=\Delta^{-1}.
\end{equation}

Fix a past up to time $i-1$ and consider two possible values $x,x'$ for $X_i$. Couple the two conditional futures successively, using an optimal Wasserstein coupling of the next-step kernels given the two pasts already constructed. Such couplings can be selected measurably on Polish spaces; see \citet{Villani2009}. If $Y_i=x$, $Z_i=x'$ and $d_\ell=\EE\rho(Y_{i+\ell},Z_{i+\ell})$, then
\[
 d_0=\rho(x,x'),\qquad d_\ell\le\sum_{j=1}^\ell\delta_jd_{\ell-j}.
\]
Induction gives
\begin{equation}\label{eq:singleperturbation}
 \EE\rho(Y_{i+\ell},Z_{i+\ell})\le r_\ell\rho(x,x').
\end{equation}
We also need to control the effect of observations before time $1$. Put $q_m=1$ for $m\le0$ and
\[
 q_\ell=\sum_{j\ge1}\delta_jq_{\ell-j}\quad(\ell\ge1).
\]
Then $0\le q_\ell\le1$ and $q_\ell\to0$. Indeed, if $b=\limsup_\ell q_\ell$, splitting the sum at a fixed $J$ gives
\[
 b\le b\sum_{j=1}^J\delta_j+\sum_{j>J}\delta_j;
\]
letting $J\to\infty$ yields $b\le ab$, hence $b=0$.
The same successive coupling argument, now starting from two arbitrary pasts at time $-N$, bounds the expected distance at time $k\ge1$ by $\Delta_Eq_{N+k}$. Consequently the range, over all initial pasts, of the corresponding conditional expectation of $F\equiv F_n$ has length at most
\begin{equation}\label{eq:forgetF}
 \Delta_E\sum_{k=1}^n c_kq_{N+k}\longrightarrow0.
\end{equation}

With these facts in mind, the rest of the proof uses a standard margingale difference argument.
Let $\cF_i=\sigma(X_t:t\le i)$ and $M_i=\EE[F\mid\cF_i]$ for $i\le n$. Extend $c_i$ by zero outside $\{1,\ldots,n\}$. Given $\cF_{i-1}$, write $M_i=G_i(X_i)$. By \eqref{eq:singleperturbation},
\[
 |G_i(x)-G_i(x')|\le\rho(x,x')\sum_{\ell\ge0}r_\ell c_{i+\ell}.
\]
Thus, conditionally on $\cF_{i-1}$, $M_i-M_{i-1}$ is centered and lies in an interval of length at most
\[
 \ell_i=\Delta_E\sum_{\ell\ge0}r_\ell c_{i+\ell}.
\]
Hoeffding's lemma \citep{Hoeffding1963} yields
\[
 \EE\bigl[e^{s(M_i-M_{i-1})}\mid\cF_{i-1}\bigr]
 \le e^{s^2\ell_i^2/8}.
\]
Iterating from $-N+1$ to $n$, and using $M_n=F$, gives
\begin{equation}\label{eq:finitepastmgf}
 \EE e^{s(F-M_{-N})}\le
 \exp\left\{\frac{s^2}{8}\sum_{i=-N+1}^n\ell_i^2\right\}.
\end{equation}
Young's convolution inequality and \eqref{eq:renewalmass} imply
\[
 \sum_{i\in\ZZ}\ell_i^2
 \le\Delta_E^2\left(\sum_{\ell \ge0}r_\ell \right)^2\sum_{k=1}^n c_k^2
 =\frac{\Delta_E^2}{\Delta^2}\sum_{k=1}^n c_k^2.
\]
By \eqref{eq:forgetF}, $|M_{-N}-\EE F|\le\Delta_E\sum_kc_kq_{N+k}$ almost surely. Since $F$ is bounded, letting $N\to\infty$ in \eqref{eq:finitepastmgf} proves \eqref{eq:mgf}. Chernoff's bound with $s=4t\Delta^2/(\Delta_E^2\sum_kc_k^2)$, followed by the same argument for $-F$ and a union bound, gives
\[
 \PP(|F-\EE F|>t)\le
 2\exp\left\{-\frac{2t^2\Delta^2}{\Delta_E^2\sum_kc_k^2}\right\}.
\]
Solving for $t$ proves \eqref{eq:oracle}.
\end{proof}

\begin{proof}[Proof of Proposition~\ref{prop:safety}]
Apply \eqref{eq:fixedaccuracy} with $\varepsilon=\kappa\Delta/2$. On the resulting event,
\[
 \Delta/2\le(1-\kappa/2)\Delta\le\widetilde\Delta_n
 \le(1+\kappa/2)\Delta\le(1+\kappa)\Delta.
\]
Division by $1+\kappa$ proves \eqref{eq:margincoverage}. A union bound with \eqref{eq:oracle} proves the empirical bound. The almost-sure convergence yields
$\Delta/\underline\Delta_n\to1+\kappa$, proving the last assertion.
\end{proof}

\subsection{Proofs of Section~\ref{sec:discrete} (the discrete case)}

We start with two lemmas. The first one shows that, when $C_n(u)$ is large enough, $\widehat p_K(\cdot\mid u)$ is close enough to some $\overline P_n(\cdot\mid u)$, which itself will be show to be close to $P_z$ for any infinite extension $z$ of $u$ in the proof or Theorem~\ref{thm:discretebound}. The second lemma shows that, with large probability, $C_n(u)$ is indeed large enough.

\begin{lemma}\label{lem:transitioncounts}
For fixed $K$, let $P_i=\mathcal L(X_i\mid\cF_{i-1})$ and, when $C_n(u)>0$, set
\[
 \overline P_n(\cdot\mid u)=\frac1{C_n(u)}
        \sum_{i=K+1}^n\ind_{\{U_i=u\}}P_i(\cdot).
\]
With probability at least $1-\eta$, simultaneously for every observed context,
\begin{equation}\label{eq:transitioncounts}
 \norm{\widehat p_K(\cdot\mid u)-\overline P_n(\cdot\mid u)}_{\TV}
 \le\sqrt{\frac{H_{n,K}(\eta)}{2C_n(u)}}.
\end{equation}
\end{lemma}
\begin{proof}[Proof of Lemma~\ref{lem:transitioncounts}]
For $u\in E^K$ and $A\subseteq E$, put
\[
 S(u,A)=\sum_{i=K+1}^n\ind_{\{U_i=u\}}
                 [\ind_{\{X_i\in A\}}-P_i(A)].
\]
The indicator of $\{U_i=u\}$ is $\cF_{i-1}$-measurable. Conditional Hoeffding and iteration give, for each fixed $\lambda\in\RR$,
\[
 \EE\exp\{\lambda S(u,A)-\lambda^2C_n(u)/8\}\le1.
\]
Write $m=n-K$ and $H=H_{n,K}(\eta)$. For each integer $1\le v\le m$, take $\lambda=\sqrt{8H/v}$. On the event $C_n(u)=v$ and $S(u,A)>\sqrt{vH/2}$, the exponent is larger than $H$. Markov's inequality, and then the same argument with $-\lambda$, give
\[
 \PP\bigl(C_n(u)=v,\ |S(u,A)|>\sqrt{vH/2}\bigr)\le2e^{-H}.
\]
A union bound over $v$, $u$, and the $2^M$ subsets $A$ has total cost $2^{M+1}M^Km e^{-H}=\eta$. On the complementary event, divide by the positive count $C_n(u)$ and take the supremum over $A$ to get~\eqref{eq:transitioncounts}.
\end{proof}

We remind that  $m=n-K$ from~\eqref{eq:notcumbersome}.
\begin{lemma}\label{lem:visits}
Under Assumption~\ref{ass:nonnull},
\[
 \PP\left(\min_{u\in E^K}C_n(u)<\frac{m\pi_*^K}{2}\right)
 \le\mathcal R_{n,K}(P).
\]
\end{lemma}
\begin{proof}[Proof of Lemma~\ref{lem:visits}]
Define $p_*=\pi_*^K$ and
$\mathcal F_t=\sigma(X_s:s\le t)$.
Fix a context $u=(u_{-1},\ldots,u_{-K})\in E^K$.
Recall that
\[
U_i=(X_{i-1},\ldots,X_{i-K}),
\qquad
C_n(u)=\sum_{i=K+1}^n \mathbf 1_{\{U_i=u\}}.
\]

First, Assumption~\ref{ass:nonnull} implies, almost surely,
\begin{equation}
\label{eq:proof:lem:visits:pstar}
\mathbb P(U_i=u\mid\mathcal F_{i-K-1})\ge \pi_*^K=:p_*.
\end{equation}

The contexts $U_{K+1},\ldots,U_n$ overlap, so we divide their indices
into $K$ groups:
\[
I_r=\{K+r,\;2K+r,\;3K+r,\ldots\}\cap\{K+1,\ldots,n\},
\qquad r=1,\ldots,K.
\]
This construction ensures that
if $i$ and $j$ are two indices in the same set $I_r$, $U_i$ and $U_j$ do not overlap.
Write $v_r=|I_r|$ and enumerate the indices in $I_r$ as follows:
$I_r = \{i_{r,\ell}=r+\ell K\text{, }\ell=1,\ldots,v_r\} $.
Note that $ v_r\ge \left\lfloor m/K \right\rfloor$, and that $\sum_{r=1}^K v_r=m$.
For each group, define
\[
Y_{r,\ell}=\mathbf 1_{\{U_{i_{r,\ell}}=u\}} \text{ and }
S_r(u)=\sum_{\ell=1}^{v_r}Y_{r,\ell},
\]
note that $C_n(u)=\sum_{r=1}^K S_r(u)$.

A direct application of~\eqref{eq:proof:lem:visits:pstar} gives:
\[
\mathbb E[Y_{r,\ell}\mid \mathcal{F}_{i_{r,\ell}-K-1}]\ge p_*
\]
and thus, for every $\lambda>0$,
\begin{align}
\nonumber
\mathbb E[e^{-\lambda Y_{r,\ell}}
          \mid\mathcal \mathcal{F}_{i_{r,\ell}-K-1}]
&=
1-(1-e^{-\lambda})
  \mathbb E[Y_{r,\ell}\mid\mathcal \mathcal{F}_{i_{r,\ell}-K-1}]
\\
\label{eq:proof:lem:visits:iter}
&\le 1-p_*+p_*e^{-\lambda}.
\end{align}
Moreover, by definition, for all $j$, $ Y_{r,j}$ is $\mathcal{F}_{i_{r,j}-1}$-measurable. So, $\sum_{j=1}^{\ell-1}Y_{r,j}$ is $\mathcal{F}_{i_{r,\ell-1}-1}$-measurable, that is, $\mathcal{F}_{i_{r,\ell}-K-1}$-measurable. Thus,
\begin{align*}
\mathbb E\exp\left(-\lambda\sum_{j=1}^{\ell}Y_{r,j}\right)
&
= \mathbb E \left\{
   \mathbb E \left[ \exp\left(-\lambda\sum_{j=1}^{\ell}Y_{r,j}\right) \mid
   \mathcal{F}_{i_{r,\ell}-K-1}
      \right]\right\}
\\
&
= \mathbb E \left\{ \exp\left(-\lambda Y_{r,\ell}\right)
   \mathbb E \left[ \exp\left(-\lambda\sum_{j=1}^{\ell-1}Y_{r,j}\right) \mid
   \mathcal{F}_{i_{r,\ell}-K-1}
      \right]\right\}
\\
& \le
(1-p_*+p_*e^{-\lambda})
\mathbb E\exp\left(-\lambda\sum_{j=1}^{\ell-1}Y_{r,j}\right)
\end{align*}
where we used~\eqref{eq:proof:lem:visits:iter} in the last line. Taking $\ell=v_r$, iterating and using $1-x\le e^{-x}$,
\[
\mathbb E[e^{-\lambda S_r(u)}]
\le (1-p_*+p_*e^{-\lambda})^{v_r}
\le \exp\{-v_rp_*(1-e^{-\lambda})\}.
\]

By Markov's inequality,
\begin{align*}
\mathbb P\left(S_r(u)<\frac{v_rp_*}{2}\right)
&\le
e^{\lambda v_rp_*/2}\,
\mathbb E[e^{-\lambda S_r(u)}]
\\
&\le
\exp\left\{
-v_rp_*\left(1-e^{-\lambda}-\frac{\lambda}{2}\right)
\right\}.
\end{align*}
Taking $\lambda=1$ and using
$1-e^{-1}-1/2>1/8$, we obtain
\[
\mathbb P\left(S_r(u)<\frac{v_rp_*}{2}\right)
\le
\exp\left(-\frac{v_rp_*}{8}\right)
\le
\exp\left(-\frac{\lfloor m/K\rfloor p_*}{8}\right).
\]
On the other hand, if $S_r(u)\ge v_rp_*/2$ for every $r$, then
\[
C_n(u)=\sum_{r=1}^K S_r(u)
\ge \frac{p_*}{2}\sum_{r=1}^K v_r
=\frac{mp_*}{2} = \frac{(n-K)\pi_*^K}{2}.
\]
Consequently, a union bound over the $K$ groups and the $M^K$
possible contexts gives
\begin{align*}
\mathbb P\left(
 \min_{u\in E^K}C_n(u)<\frac{(n-K)\pi_*^K}{2}
\right)
&\le
\sum_{u\in E^K}\sum_{r=1}^K
\mathbb P\left(S_r(u)<\frac{v_rp_*}{2}\right)
\\
&\le
KM^K
\exp\left\{
-\frac{\lfloor(n-K)/K\rfloor\pi_*^K}{8}
\right\}
\\
&=\mathcal R_{n,K}(P)
\end{align*}
which is exactly the statement of the lemma.
\end{proof}

Equipped with Lemmas~\ref{lem:transitioncounts} and~\ref{lem:visits}, we are ready to prove Theorem~\ref{thm:discretebound} and its corollary.
\begin{proof}[Proof of Theorem~\ref{thm:discretebound}]
Note that $  \mathbf{1}_{\{U_i=u\}} P_i$ can be rewritten as $P_{Z}$ for some (random) infinite extension $Z$ of $u$. Thus, for any other infinite extension $z$ of $u$,
$$  \norm{ P_z - \mathbf{1}_{\{U_i=i\}} P_i }_{\TV} = \norm{ P_z - P_{Z} }_{\TV} \leq T_K. $$
Thus,
$$  \norm{ P_z - \overline P_n(\cdot\mid u) }_{\TV} \leq T_K. $$
Moreover, on the event of Lemma~\ref{lem:transitioncounts},
$$
 \norm{\widehat p_K(\cdot\mid u)-\overline P_n(\cdot\mid u)}_{\TV}
 \le\sqrt{\frac{H_{n,K}(\eta)}{2C_n(u)}} \leq \max_{u\in E^K} \sqrt{\frac{H_{n,K}(\eta)}{2C_n(u)}}  =  r_{n,K}(\eta).
$$
The triangle inequality gives:
\begin{equation}\label{eq:discreteextension}
 \norm{\widehat p_K(\cdot\mid u)-P_z}_{\TV}\le r_{n,K}(\eta) +T_K.
\end{equation}

This gives us all the required tools to control $ |\widehat\delta_{j,K}-\delta_j|$. In order to do so, fix $j\le K$ and $\alpha\in E$, and define $z^{(j,\alpha)}$ such that $z^{(j,\alpha)}_i=z_i$ for aall $i\neq j$, and  $z^{(j,\alpha)}_j = \alpha $, and $u^{(j,\alpha)}$ the initial segment of $z^{(j,\alpha)}$. Thus
\begin{align*}
 \norm{\widehat p_K(\cdot\mid u)-\widehat p_K(\cdot\mid u^{(j,\alpha)})}_{\TV} & -  \norm{P_{z}-P_{z^{(j,\alpha)}}}_{\TV}
\\
& \le
  \norm{\widehat p_K(\cdot\mid u)-\widehat p_K(\cdot\mid u^{(j,\alpha)}) - P_z +P_{z^{(j,\alpha)}} }_{\TV}
  \\
 & \le \norm{\widehat p_K(\cdot\mid u)-P_z}_{\TV} + \norm{\widehat p_K(\cdot\mid u^{(j,\alpha)})-P_{z^{(j,\alpha)}}}_{\TV}
 \\
 & \le 2 \left( r_{n,K}(\eta) +T_K \right)
\end{align*}
where we used~\eqref{eq:discreteextension} in the last line. Put
 $$ \alpha_0 = \arg\max_{\alpha\in E} \norm{\widehat p_K(\cdot\mid u)-\widehat p_K(\cdot\mid u^{(j,\alpha)})}_{\TV}, $$
we have
$$ \widehat\delta_{j,K} \leq  \norm{P_{z}-P_{z^{(j,\alpha_0)}}}_{\TV} + 2 \left( r_{n,K}(\eta) +T_K \right) \leq \delta_j +  2 \left( r_{n,K}(\eta) +T_K \right)  .$$
A symmetric argument gives:
$$
\norm{P_{z}-P_{z^{(j,\alpha)}  }}_{\TV}
-
 \norm{\widehat p_K(\cdot\mid u)-\widehat p_K(\cdot\mid u^{(j,\alpha)}   )}_{\TV}
 \leq 2 \left( r_{n,K}(\eta) +T_K \right),
$$
this time take
$$
\alpha_1 = \arg\max_{\alpha\in E} \norm{P_{z}-P_{z'}}_{\TV}
$$
to get:
$$ \delta_j \leq   \norm{\widehat p_K(\cdot\mid u)-\widehat p_K(\cdot\mid u^{(j,\alpha_1)}   )}_{\TV} + 2 \left( r_{n,K}(\eta) +T_K \right) \leq \widehat\delta_{j,K} +  2 \left( r_{n,K}(\eta) +T_K \right)  .$$
Thus,
\begin{equation}\label{eq:deltacomparison}
 |\widehat\delta_{j,K}-\delta_j|\le 2 \left( r_{n,K}(\eta) +T_K \right).
\end{equation}

We can now derive the upper bound on the estimation error of $\widetilde{\Delta}_{n,K}$:
\begin{align*}
\left|
\widetilde{\Delta}_{n,K}
- \Delta
\right|
& = 
\left| \clip\left(1-\sum_{j=1}^K\widehat\delta_{j,K}\right) - \Delta \right|
\\
&
\leq
\left| \left(1-\sum_{j=1}^K\widehat\delta_{j,K}\right) - \Delta \right|
\\
& =
\left| \sum_{j=1}^K\widehat\delta_{j,K} - \sum_{j=1}^\infty \delta_j \right|
\\
&
\leq
\sum_{j=1}^K  |\widehat\delta_{j,K}-\delta_j| + T_K
\\
& \leq \sum_{j=1}^K  2 \left( r_{n,K}(\eta) +T_K \right)+ T_K
\end{align*}
from~\eqref{eq:deltacomparison}. Thus,
$$
\left|
\widetilde{\Delta}_{n,K}
- \Delta
\right|
\leq (2K+1) T_K + 2 \sum_{j=1}^K r_{n,K}(\eta),
$$
that is~\eqref{eq:discreteobserved}. Lemma~\ref{lem:visits} implies
$r\le\sqrt{H_{n,K}(\eta)/(m\pi_*^K)}$ outside an event of probability $\mathcal R_{n,K}(P)$. A union bound gives~\eqref{eq:discretedeterministic}.
\end{proof}

\begin{proof}[Proof of Corollary~\ref{cor:discrete}]
Fix $q>0$ and apply Theorem~\ref{thm:discretebound} with $\eta=n^{-(q+2)}$. As $K_n=o(\log n)$ and $\pi_*>0$ is a constant,
\[
 \pi_*^{K_n}=n^{-o(1)},\qquad M^{K_n}=n^{o(1)},\qquad
 H_{n,K_n}(n^{-(q+2)})=O_{P,q}(\log n).
\]
The stochastic term in \eqref{eq:discretedeterministic} is $n^{-1/2+o(1)}$ and tends to zero. Moreover,
\[
 \mathcal R_{n,K_n}(P)
 =\exp\{-n^{1-o(1)}+o(\log n)\}
\]
and is eventually smaller than $n^{-(q+2)}$. The deterministic bias tends to zero by $K_nT_{K_n}\to0$. For every fixed $\varepsilon>0$, these facts imply \eqref{eq:fixedaccuracy}, since $2n^{-(q+2)}\le n^{-q}$ eventually. Taking a summable failure level and applying Borel--Cantelli proves almost-sure convergence.
\end{proof}

\subsection{Proofs of Section~\ref{sec:continuous} (the continuous case)}

Throughout this subsection, a constant denoted by $A_P$ may be increased from one display to the next, but is independent of $n,K,h,\eta$.

We first prove  Proposition~\ref{prop:eq:densityW}.
\begin{proof}[Proof of Proposition~\ref{prop:eq:densityW}]
We refer the reader to~\citet[Chapter 8 and 9]{Brezis2011} for Neumann-Poincaré inequalities and Sobolev spaces $W^{1,p}(E)$ on domains $E$. These inequalities take the form: for any $\varphi \in W^{1,p}(E)$,
\[
 \norm{\varphi-\bar\varphi}_2\le C_E \norm{\nabla\varphi}_2 
\]
where $\bar\varphi:=\int_{E}\varphi(x)\,{\rm d}x$.
From~\cite{PayneWeinberger1960}, we know that Neumann-Poincaré inequalities is satisfied on the unit interval $E=[0,1]$ with constant $C_E = 1/\pi$. By tensorization, for $E=[0,1]^d$ we also have $C_E = 1/\pi$. Let $\lambda$ denote the Lebesgue measure on $[0,1]^d$.
Using Kantorovich--Rubinstein duality \citep{Villani2009}, for two probability distributions $P$ and $Q$ with densities ${\rm d}P/{\rm d}\lambda = p$ and ${\rm d}Q/{\rm d}\lambda = q$ ,
\begin{align*}
 W_{1,2}( P, Q )
 & = \sup_{\Lip(\varphi)\le 1}\int \varphi (p-q) {\rm d}\lambda
 \\
 &=\sup_{\Lip(\varphi)\le 1}\int(\varphi-\bar\varphi)(p-q) {\rm d}\lambda \text{ as } \int(p-q){\rm d}\lambda =0,
 \\
 & \le \sup_{\Lip(\varphi)\le 1} \norm{\varphi-\bar\varphi}_2 \norm{p-q}_\infty \text{ by Cauchy-Schwarz},
 \\
 &\le \sup_{\Lip(\varphi)\le 1} \pi^{-1}\norm{\nabla\varphi}_2 \norm{p-q}_\infty \text{ by Neumann-Poincaré},
 \\
 & \le\pi^{-1}\norm{p-q}_\infty \text{ as } \norm{\nabla\varphi}_2\le1 \text{  because the cube has volume } 1.
\end{align*}
Take $P = P_u $ and $Q=P_v$ to conclude.
\end{proof}

We now turn to the proof of Theorem~\ref{thm:continuousbound} (and its corollary). As in the previous subsections, we begin with a series of lemmas.

\begin{lemma}
\label{lem:blockregularity}
Let $g_K$ be the density of $(X_{-1},\ldots,X_{-K})$, $f_K$ the density of $(X_0,X_{-1},\ldots,X_{-K})$, and $p_K=f_K/g_K$. Under Assumption~\ref{ass:continuous},
\begin{equation}\label{eq:blockbounds}
 \forall u\in E^K\text{, }
 g_K(u)\ge\alpha^K\text{, and }
 \max_{a\in\{f_K,g_K,p_K\}}\ \max_{|\beta|\le2}
                  \norm{\partial^\beta a}_\infty\le A_P^{K+1}
\end{equation}
for a constant $A_P\ge2$.
\end{lemma}
\begin{proof}[Proof of Lemma~\ref{lem:blockregularity}]
Let $\zeta=(X_{-K-1},X_{-K-2},\ldots)$, with marginal law $\mu_K$. Conditional on $\zeta$, the density of the block $u\in E^K$ is
\begin{equation}\label{eq:blocklikelihood}
 G_K(u,\zeta)=\prod_{r=1}^K
       p(u_{-r}\mid u_{-r-1},\ldots,u_{-K},\zeta).
\end{equation}
The most ancient factor is the one with $r=K$. Thus
\[
 g_K(u)=\int G_K(u,\zeta)\,\mu_K({\rm d}\zeta),\qquad
 f_K(x,u)=\int p(x\mid u,\zeta)G_K(u,\zeta)\,\mu_K({\rm d}\zeta).
\]
The lower bound on $g_K$ follows directly from the lower bound on $p(u_{-r}|z)\geq \alpha$ for any $z$. It also gives full support for the density of the infinite past $f_K(x,u)$, since every finite open cylinder has positive probability. Two continuous density-kernel versions compatible with the same stationary law must therefore coincide everywhere. The kernel is bounded above by a finite number $B$, by continuity on the compact space $E\times\past$. The first derivative of a product of $K+1$ factors is a sum of at most $K+1$ products; the second derivative is a sum of at most $(K+1)^2$ products. Thanks to Assumption~\ref{ass:continuous}, all the differentiated factors in these products are bounded by $H$, and the others by $B$. Differentiation under the integral is therefore justified, and we also have the continuity of the derivatives by dominated convergence. All these bounds are at most $A_P^{K+1}$ for some constant $A_P$. The quotient rule, $g_K\ge\alpha^K$, and an increase of $A_P$ give the same conclusion for $p_K$.
\end{proof}

\begin{lemma}
\label{lem:feedback}
Put $Q=\frac{L_0+L}{\alpha}$ for short.
For $j\le K$, write
\[
 L_j^{(K)}=\sup_{x\in E,u\in E^K}\norm{\nabla_{u_{-j}}\,p_K(x\mid u)}_2.
\]
Then
\begin{equation}\label{eq:feedbackbound}
 |L_j^{(K)}-L_j|\le\omega_{j,K}+2\sqrt d\,Q T_K^L,
 \qquad
 \abs{\sum_{j=1}^K L_j^{(K)}-L}\le B_K
\end{equation}
where we remind that $B_K$ is given by~\eqref{eq:continuousbias}.
\end{lemma}
\begin{proof}[Proof of Lemma~\ref{lem:feedback}]
For $u\in E^K$, a version of the density in $\zeta$ of the conditional law of the remote past given $u$ is given by
\[
 \nu_{K,u}(d\zeta)=\frac{G_K(u,\zeta)}{g_K(u)}\,\mu_K(d\zeta).
\]
For a unit vector $v\in\RR^d$, let $\partial_{j,v}$ denote the differentiation with respect to $u_{-j}$ in direction $v$. Let $S_{j,v}(u,\zeta):=\partial_{j,v}\log G_K(u,\zeta)$ denote the conditional score function. From~\eqref{eq:blocklikelihood},
\begin{equation}
\label{eq:scorebound}
 |S_{j,v}(u,\zeta)|\le
 \frac{L_0+\sum_{l=1}^{j-1}L_l}{\alpha}\le Q.
\end{equation}
Differentiating the density of $\nu_{K,u}$ gives
\[
 \partial_{j,v}\nu_{K,u}(d\zeta)
 =\left[S_{j,v}(u,\zeta)-\int S_{j,v}(u,\xi )\nu_{K,u}({\rm d}\xi)\right]
      \nu_{K,u}(d\zeta).
\]
This is a signed measure which satisfies
\begin{equation}
\label{eq:centerandTV}
 \int \partial_{j,v}\nu_{K,u}({\rm d}\zeta) = 0
 \text{ and }
 \int \left| \partial_{j,v}\nu_{K,u} \right| ({\rm d}\zeta) \leq 2Q.
\end{equation}
Now, let us turn to $p_K(x\mid u)$. We actually have $p_K(x\mid u)=\int p(x\mid u,\zeta)\nu_{K,u}({\rm d}\zeta)$, and thus:
\begin{equation*}
 \partial_{j,v}p_K(x\mid u)
 =\int\partial_{j,v}p(x\mid u,\zeta)\nu_{K,u}({\rm d}\zeta)
  +\int p(x\mid u,\zeta)\partial_{j,v}\nu_{K,u}({\rm d}zeta).
\end{equation*}
For any fixed extension $\zeta'$ of $u$, we thus have
\begin{multline}
\label{eq:feedbackidentity}
 \partial_{j,v}p_K(x\mid u)-\partial_{j,v}p_K(x\mid u,\zeta')
 =\int \partial_{j,v}p(x\mid u,\zeta)\nu_{K,u}({\rm d}\zeta)-\partial_{j,v}p_K(x\mid u,\zeta')
  \\
  +\int p(x\mid u,\zeta)\partial_{j,v}\nu_{K,u}({\rm d}\zeta).
\end{multline}
We first control the last term the r.h.s. in~\eqref{eq:feedbackidentity}.
Thanks to~\eqref{eq:centerandTV}, we have
$$
\int p(x\mid u,\zeta)\partial_{j,v}\nu_{K,u}({\rm d}\zeta) = 
\int [p(x\mid u,\zeta)-p(x\mid u,\zeta')]\partial_{j,v}\nu_{K,u}({\rm d}\zeta)
$$
and thus
\begin{align*}
\int\partial_{j,v}p(x\mid u,\zeta)\nu_{K,u}({\rm d}\zeta)
& \le  \int \left|p(x\mid u,\zeta)-p(x\mid u,\zeta')\right| \left|\partial_{j,v}\nu_{K,u}\right|({\rm d}\zeta)
\\
& \le \sup_{\zeta} \int \left|p(x\mid u,\zeta)-p(x\mid u,\zeta')\right| \int \left|\partial_{j,v}\nu_{K,u}\right|({\rm d}\zeta)
\\
& \leq 2Q \sqrt{d} \sum_{\ell>K} L_{\ell} \text{ from~\eqref{eq:centerandTV} and~\eqref{eq:densitylip}}
\\
& = 2Q \sqrt{d} T_K^L.
\end{align*}
Coming back to~\eqref{eq:feedbackidentity}, we now control the first two terms in the r.h.s:
\begin{align*}
\int \partial_{j,v}p(x\mid u,\zeta)\nu_{K,u}({\rm d}\zeta)-\partial_{j,v}p_K(x\mid u,\zeta')
& =
\int \left[ \partial_{j,v}p(x\mid u,\zeta) - \partial_{j,v}p_K(x\mid u,\zeta') \right]\nu_{K,u}({\rm d}\zeta)
\\
& \le \int  \omega_{j,K} \nu_{K,u}({\rm d}\zeta) = \omega_{j,K}.
\end{align*}
Plugging these upper bounds in \eqref{eq:feedbackidentity} and taking the supremum over the direction $v$ gives:
\[
 \norm{\nabla_jp_K(x\mid u)-\nabla_jp(x\mid u,\zeta')}_2
 \le\omega_{j,K}+2Q\sqrt d\,T_K^L.
\]
This gives the first equation in~\eqref{eq:feedbackbound}. The second one is a direct consequence:
\begin{align*}
\abs{\sum_{j=1}^K L_j^{(K)}-L}
& \leq \sum_{j=1}^K \abs{L_j^{(K)}-\sum_{j=1}^K L_j} +  \sum_{j=K+1}^\infty L_j
\\
& \leq \sum_{j=1}^K \left(\omega_{j,K}+2Q\sqrt d\,T_K^L \right) + T_{K}^L
\\
& = \sum_{j=1}^K \omega_{j,K}+\left(1+2KQ\sqrt d\right)T_K^L
\\
& = \Omega_{j,K} + \left(1+\frac{2K \sqrt d (L_0+L)}{\alpha}\right)T_K^L = B_K.
\end{align*}
The proof is complete.
\end{proof}

\begin{lemma}
\label{lem:kernel}
Let $\mathcal A_K$ consist of $f_K,g_K$, and all their first scalar derivatives. For $a\in \mathcal A_K$, if $a\in\{f_K,g_K\}$, let $\widehat a_h$ be its corresponding estimator given by~\eqref{eq:kernel:estim:1} and~\eqref{eq:kernel:estim:2}; if $a$ is a derivative of such a function, let $\widehat a_h$ be the derivative of the kernel estimator. Under Assumption~\ref{ass:continuous}, for $n\ge2K+2$, $0<h\le1/4$, and any $0<\eta<1$,
\begin{equation}\label{eq:primitiveerror}
 \PP\left(\max_{a\in\mathcal A_K}
       \norm{\widehat a_h-a}_{\infty,\mathcal I_{K,h}}
 \le A_P^{K+1}\left[h+h^{-(D+2)}
                 \sqrt{\frac{V_{n,K,h}(\eta)}n}\right]\right)\ge1-\eta.
\end{equation}
We remind $D=d(K+1)$ from~\eqref{eq:notcumbersome}.
\end{lemma}
\begin{proof}[Proof of Lemma~\ref{lem:kernel}]
Put
\[
W_i=(X_i,X_{i-1},\ldots,X_{i-K})\in[0,1]^D,
\qquad i=K+1,\ldots,n.
\]
We distinguish throughout between the evaluation point
$z=(x,u)\in I_{K,h}$ and the observations entering $W_i$.
We first control the fluctuations of $\widehat{a}_h(z)$ around its expectation $E\widehat a_h(z)$ at a fixed evaluation point $z$(variance term). This will be done by using Theorem~\ref{thm:concentration}. In a first step, we check that $F_n=F(X_1,\dots,X_n):=\widehat{a}_h(z)$ satisfies the assumptions of Theorem~\ref{thm:concentration}. In the second step, we can thus apply the theorem and get the control of $\widehat{a}_h(z)-E\widehat a_h(z)$; we make this control uniform on $a\in\mathcal{A}_K$ through union bounds. The third step provides a bound on the bias.

\medskip
\noindent
\textit{Step 1.}
Assume that  $
c_k\ge
\max\bigl\{2,\|k\|_\infty,\|k'\|_\infty,\|k''\|_\infty\bigr\}.
$
For $z,w\in\mathbb R^D$, define
\[
\Phi_h(z,w)
=
h^{-D}\prod_{\ell=1}^D
k\left(\frac{w_\ell-z_\ell}{h}\right).
\]
Then
\[
\widehat f_{K,h}(z)
=
\frac1m\sum_{i=K+1}^n\Phi_h(z,W_i).
\]
Fix $r\in\{1,\ldots,D\}$. Differentiating with respect to the
evaluation coordinate $z_r$ gives
\[
\partial_{z_r}\widehat f_{K,h}(z)
=
\frac1m\sum_{i=K+1}^n\Psi_{r,h}(z,W_i),
\]
where
\[
\Psi_{r,h}(z,w)
:=
\partial_{z_r}\Phi_h(z,w)
=
-h^{-(D+1)}k'\left(\frac{w_r-z_r}{h}\right)
\prod_{\ell\ne r}k\left(\frac{w_\ell-z_\ell}{h}\right).
\]Since this
product contains $D$ factors each bounded by $c_k$, we have
\[
|\Psi_{r,h}(z,w)|
\le c_k^D h^{-(D+1)}.
\]

To apply Theorem~\ref{thm:concentration}, we need to know how this summand changes when
one observation changes. We therefore differentiate once more, this
time with respect to a data coordinate $w_s$:
\[
\partial_{w_s}\Psi_{r,h}(z,w)
=
-h^{-(D+2)}
\begin{cases}
k''\left(\frac{w_r-z_r}{h}\right)\displaystyle\prod_{\ell\ne r}k\left(\frac{w_\ell-z_\ell}{h}\right),
& s=r,\\[2mm]
k'\left(\frac{w_r-z_r}{h}\right)k'\left(\frac{w_s-z_s}{h}\right)
\displaystyle\prod_{\ell\notin\{r,s\}}k\left(\frac{w_\ell-z_\ell}{h}\right),
& s\ne r.
\end{cases}
\]
Both cases give
\[
|\partial_{w_s}\Psi_{r,h}(z,w)|
\le c_k^D h^{-(D+2)}.
\]
Notice that a single observation is a vector in $[0,1]^d$.
Changing one observation within a block therefore changes 
$d$ of the $D$ scalar coordinates of that block. The Euclidean norm
of the gradient with respect to this $d$-vector is thus bounded by
\[
\sqrt d\,c_k^D h^{-(D+2)}.
\]
Consequently, if $w$ and $w'$ differ only in one such $d$-vector,
say $w^{(j)}$ and $w'^{(j)}$, the mean value inequality gives
\[
|\Psi_{r,h}(z,w)-\Psi_{r,h}(z,w')|
\le
\sqrt d\,c_k^D h^{-(D+2)}
\|w^{(j)}-w'^{(j)}\|_2.
\]

We now pass from one summand to the whole empirical average.
If the observation $X_s$ is replaced by $X'_s$, then $W_i$ changes
only when
$
i-K\le s\le i$, or equivalently, $s\le i\le s+K$.
Thus at most $K+1$ summands are affected. After dividing by
$m=n-K$, we obtain
\begin{multline*}
\left|
\partial_{z_r}\widehat f_{K,h}(z;X_{1:n})
-
\partial_{z_r}\widehat f_{K,h}
(z;X_1,\ldots,X_{s-1},X'_s,X_{s+1},\ldots,X_n)
\right|
\\
\le
\underbrace{\frac{\sqrt d\,(K+1)c_k^D}{n-K}\,
h^{-(D+2)}}_{=: b_{n,K,h}} \|X_s-X'_s\|_2.
\end{multline*}

The same upper bound applies to the undifferentiated density
estimate: its summands require only one differentiation with
respect to the observations, giving $h^{-(D+1)}$, which is bounded
by $h^{-(D+2)}$ because $h\le1$.
It also applies to $\widehat g_{K,h}$ and its first derivatives: as their dimension is $dK\le D$, and changing one observation affects at most $K$ summands.

This shows that $F(X_1,\dots,X_n)$ satisfies the assumptions of Theorem~\ref{thm:concentration}.

\medskip
\noindent
\textit{Step 2.} As announced, we now apply Theorem~\ref{thm:concentration}. As $\Delta_E=\sqrt d$, we obtain, for every $t>0$,
\begin{align*}
\mathbb P\left(
|\widehat a_h(z)-\mathbb E\widehat a_h(z)|>t
\right)
&\le
2\exp\left\{
-\frac{2\Delta_L^2t^2}{d\,n b_{n,K,h}^2}
\right\}\\
&=
2\exp\left\{
-\frac{
2(n-K)^2\Delta_L^2 h^{2(D+2)}t^2
}{
n d^2(K+1)^2c_k^{2D}
}
\right\}\\
&\le
2\exp\left\{
-\frac{
n\Delta_L^2 h^{2(D+2)}t^2
}{
2d^2(K+1)^2c_k^{2D}
}
\right\},
\end{align*}
where we used $n-K\ge n/2$.
Since $D=d(K+1)$, there exists $A_0\ge2$, depending only on
$d$ and the fixed kernel, such that for any $K\ge 1$,
\[
2d^2(K+1)^2c_k^{2D}\le A_0^{2(K+1)}.
\]
Thus
\begin{equation}
\label{eq:pointwise:concentration:proof}
\mathbb P\left(
|\widehat a_h(z)-\mathbb E\widehat a_h(z)|>t
\right)
\le
2\exp\left\{
-\frac{
n\Delta_L^2 h^{2(D+2)}t^2
}{
A_0^{2(K+1)}
}
\right\}.
\end{equation}
We will now make the concentration uniform in $a\in\mathcal{A}_K$ and $z$. We will discretize the centered stochastic process
\[
Z_a(z):=\widehat a_h(z)-\mathbb E\widehat a_h(z)
\]
so that we can use a (finite) union bound. This is possible thanks to the regularity of $\widehat a_h(z)$ in $z$ (as opposed to the regularity with respect to the observations used in Step~1).
The kernel calculations of Step~1 also show that, for every scalar
coordinate $s$,
\[
|\partial_{z_s}\widehat a_h(z)|
\le c_k^D h^{-(D+2)}.
\]
Indeed, differentiating with respect to $z_s$ instead of $w_s$
changes only the sign of the derivative of a function depending
on $w-z$. Taking expectations gives the same bound for
$\partial_{z_s}\mathbb E\widehat a_h(z)$.
Hence
\[
|Z_a(z)-Z_a(z')|
\le
2c_k^D h^{-(D+2)}
\sum_{s=1}^D|z_s-z'_s|.
\]
Let $\mathcal G$ be a product grid in $I_{K,h}=[h,1-h]^D$
with mesh at most
\[
\epsilon_0=
\frac{h^{D+2}}{4n(D+1)c_k^D}.
\]
We may choose it so that
\[
|\mathcal G|\le(1+2/\epsilon_0)^D,
\]
and every $z\in I_{K,h}$ has a grid point $z^\circ\in\mathcal G$
with $|z_s-z^\circ_s|\le\epsilon_0$ for all $s$. Consequently,
\[
|Z_a(z)-Z_a(z^\circ)|
\le
2D c_k^D h^{-(D+2)}\epsilon_0
=
\frac{D}{2n(D+1)}
\le\frac1n.
\]
Thus it is enough to control the centered process on the finite
grid, at the cost of an additional error $1/n$.
There are at most $2(D+1)$ elements $a\in\mathcal A_K$, and thus at most $2(D+1)$ processes$ (Z_a(z))$.
Set
\[
\Lambda=
\log\left(
\frac{4(D+1)(1+2/\epsilon_0)^D}{\eta}
\right) \text{ and }
t_*=
\frac{A_0^{K+1}}{\Delta_L}\,
h^{-(D+2)}\sqrt{\frac{\Lambda}{n}}.
\]
An application of~\eqref{eq:pointwise:concentration:proof} and a union bound over $z\in\mathcal G$ and $a\in\mathcal A_K$ gives:
\[
\mathbb P\left(
\max_{a\in\mathcal A_K}
\max_{z\in\mathcal G}|Z_a(z)|>t_*
\right)
\le
4(D+1)(1+2/\epsilon_0)^D e^{-\Lambda}
=
\eta.
\]
Therefore, with probability at least $1-\eta$,
\[
\max_{a\in\mathcal A_K}
\|Z_a\|_{\infty,I_{K,h}}
\le t_*+\frac1n.
\]
To simplify this bound, observe that $\epsilon_0\le1$ and
$
\log(1+2/\epsilon_0)
\le
\log\bigl(12n(D+1)\bigr)
+D\log c_k
+(D+2)\log(1/h)$.
Using $D=d(K+1)\le d(n+1)$, we conclude that
\[
\Lambda
\le
C_{d,k}(D+1)^2
\log\left(\frac{e(n+1)}{\eta h}\right)
=
C_{d,k}V_{n,K,h}(\eta)
\]
for a constant $C_{d,k}$ independent of $n,K,h,\eta$.
The factor $\Delta_L^{-1}$ can be absorbed into a constant
depending on $P$, and
\[
\frac1n
\le
h^{-(D+2)}
\sqrt{\frac{V_{n,K,h}(\eta)}{n}}.
\]
After increasing a constant $A_P\ge2$, we have proved
\[
\mathbb P\left(
\max_{a\in\mathcal A_K}
\|\widehat a_h-\mathbb E\widehat a_h\|_{\infty,I_{K,h}}
\le
A_P^{K+1}h^{-(D+2)}
\sqrt{\frac{V_{n,K,h}(\eta)}{n}}
\right)
\ge1-\eta.
\]

\medskip
\noindent
\textit{Step 3.}
Recall that $\mathcal K_D(v)=\prod_{\ell=1}^D k(v_\ell)$.
By stationarity, each $W_i$ has density $f_K$. Thus, for
$z\in I_{K,h}$, a change of variables gives
\begin{align*}
\mathbb E\widehat f_{K,h}(z)
&=
\int_{[0,1]^D}
h^{-D}\mathcal K_D\left(\frac{w-z}{h}\right)f_K(w)\,{\rm d}w\\
&=
\int_{[-1,1]^D}
\mathcal K_D(v)f_K(z+hv)\,{\rm d}v.
\end{align*}
The restriction $z\in[h,1-h]^D$ ensures that
$z+hv\in[0,1]^D$ throughout the support of the kernel.

For a first derivative, the explicit formula from Step~1 yields
\[
\mathbb E\partial_{z_r}\widehat f_{K,h}(z)
=
-\frac1h
\int_{[-1,1]^D}
k'(v_r)\prod_{\ell\ne r}k(v_\ell)
f_K(z+hv)\,{\rm d}v.
\]
Integrating by parts with respect to $v_r$, the boundary term
vanishes because $k(-1)=k(1)=0$. Since
\[
\partial_{v_r}\bigl[f_K(z+hv)\bigr]
=
h\,(\partial_r f_K)(z+hv),
\]
we obtain
\[
\mathbb E\partial_{z_r}\widehat f_{K,h}(z)
=
\int_{[-1,1]^D}
\mathcal K_D(v)(\partial_r f_K)(z+hv)\,{\rm d}v.
\]
The same argument applies to $g_K$ and its first derivatives in
$dK$ dimensions.
Consequently, for every $a\in\mathcal A_K$,
\[
\mathbb E\widehat a_h(z)-a(z)
=
\int_{[-1,1]^D}
\mathcal K_D(v)\bigl[a(z+hv)-a(z)\bigr]\,{\rm d}v.
\]
Lemma~\ref{lem:blockregularity} bounds all first and second scalar
derivatives of $f_K$ and $g_K$ by $A_P^{K+1}$. Thus, whether $a$ is a density or one of its first
derivatives,
\[
|a(z+hv)-a(z)|
\le
A_P^{K+1}h\sum_{\ell=1}^D|v_\ell|
\le
D A_P^{K+1}h.
\]
Because $\mathcal K_D$ is nonnegative and integrates to one,
\[
\max_{a\in\mathcal A_K}
\|\mathbb E\widehat a_h-a\|_{\infty,I_{K,h}}
\le
D B_P^{K+1}h.
\]
Finally, $D=d(K+1)$ can be absorbed into a constant exponential
in $K+1$. Combining this bias bound with Step~3 and increasing
$A_P$ once more gives
\[
\mathbb P\left(
\max_{a\in\mathcal A_K}
\|\widehat a_h-a\|_{\infty,I_{K,h}}
\le
A_P^{K+1}
\left[
h+
h^{-(D+2)}
\sqrt{\frac{V_{n,K,h}(\eta)}{n}}
\right]
\right)
\ge1-\eta,
\]
as claimed.
\end{proof}

\begin{lemma}
\label{lem:ratio}
For a sufficiently large constant $C_P$ in \eqref{eq:continuouserror}, if $e_{n,K,h}(P,\eta)<1$, then, on the event of Lemma~\ref{lem:kernel},
\begin{equation}\label{eq:Lfiniteest}
 \frac1\pi\sum_{j=1}^K|\widehat L_{j,K,h}-L_j^{(K)}|
 \le e_{n,K,h}(P,\eta).
\end{equation}
\end{lemma}
\begin{proof}[Proof of Lemma~\ref{lem:ratio}]
In all the proof, we work on the event of Lemma~\ref{lem:kernel}. Put $\beta=A_P^{K+1}t$
and
\[
t=h+h^{-(D+2)}
\sqrt{\frac{V_{n,K,h}(\eta)}{n}}
\]
so that
\[
C_P^{K+1}t=e_{n,K,h}(P,\eta)<1.
\]
If $C_P$ is chosen large enough to ensure $C_P\ge 2A_P / \alpha$, we have
\[
\beta=A_P^{K+1}t
<
\left(\frac{A_P}{C_P}\right)^{K+1}
\le
\left(\frac{\alpha}{2}\right)^{K+1}
\le \frac{\alpha^K}{2}.
\]
Moreover, $h\le1$ and $V_{n,K,h}(\eta)\ge1$ imply
$t\ge n^{-1/2}$. Therefore
\[
C_P^{K+1}n^{-1/2}<1,
\qquad\text{and hence}\qquad
n^{-1}<C_P^{-2(K+1)}
\le\frac{\alpha^K}{2}.
\]
Consequently, throughout the interior cube $[h,1-h]^{dK}$,
\begin{equation}
\label{eq:betahat:lowerbound}
\widehat g_{K,h}
\ge g_K-\beta
\ge \frac{g_K}{2}
\ge \frac{\alpha^K}{2}
\ge \frac{1}{n}.
\end{equation}
Thus
\[
\widehat g_{K,h}^{\,\tau}
=\widehat g_{K,h}\vee \frac{1}{n}
=\widehat g_{K,h}.
\]
So, in this setting, in the formula defining $\widehat D_{j,K,h}$, we may
replace the trimmed denominator by $\widehat g_{K,h}$.

With this observation in mind, we will control $\widehat D_{j,K,h}(x,u)-\nabla_jp_K(x\mid u)$ on the interior cube $I_{K,h}=[h,1-h]^{D}$.
Fix $j\in\{1,\ldots,K\}$ and a scalar coordinate
$r\in\{1,\ldots,d\}$ of $u_{-j}$. Let $\partial_{j,r}$ denote
differentiation with respect to this coordinate. Define
\[
\begin{aligned}
f(x,u)&=f_K(x,u),
&\qquad g(u)&=g_K(u),\\
a(x,u)&=\partial_{j,r}f_K(x,u),
& b(u)&=\partial_{j,r}g_K(u),
\end{aligned}
\]
and their empirical counterparts
\[
\begin{aligned}
\widehat f(x,u)&=\widehat f_{K,h}(x,u),
&\qquad \widehat g(u)&=\widehat g_{K,h}(u),\\
\widehat a(x,u)&=\partial_{j,r}\widehat f_{K,h}(x,u),
& \widehat b(u)&=\partial_{j,r}\widehat g_{K,h}(u).
\end{aligned}
\]
In particular, $g$, $b$, and their estimators depend only on $u$,
whereas $f$, $a$, and their estimators depend on both $x$ and $u$.
Fix $(x,u)\in I_{K,h}$. We have
\[
\partial_{j,r}p_K(x\mid u)
=
\frac{a(x,u)}{g(u)}
-
\frac{f(x,u)b(u)}{g(u)^2}
\]
and
\[
\bigl[\widehat D_{j,K,h}(x,u)\bigr]_r
=
\frac{\widehat a(x,u)}{\widehat g(u)}
-
\frac{\widehat f(x,u)\widehat b(u)}{\widehat g(u)^2}.
\]
Thus,
\begin{equation}
\label{eq:termstocontrol}
\partial_{j,r}p_K(x\mid u)-\bigl[\widehat D_{j,K,h}(x,u)\bigr]_r
=
\left(
\frac{a(x,u)}{g(u)}-\frac{\widehat a(x,u)}{\widehat g(u)}
\right)
+ \left(\frac{f(x,u)b(u)}{g(u)^2}-\frac{\widehat f(x,u)\widehat b(u)}{\widehat g(u)^2}
\right).
\end{equation}
As we are in the event of Lemma~\ref{lem:kernel}, all four estimation errors satisfy
\[
\begin{aligned}
|\widehat f(x,u)-f(x,u)|&\le\beta,
&\qquad |\widehat g(u)-g(u)|&\le\beta,\\
|\widehat a(x,u)-a(x,u)|&\le\beta,
& |\widehat b(u)-b(u)|&\le\beta.
\end{aligned}
\]
Moreover, by assumption, $g(u)\ge\alpha^K$, and~\eqref{eq:betahat:lowerbound} yields
\begin{equation}
\label{eq:termstocontrol2}
\beta\le\frac{\alpha^K}{2}\le\frac{g(u)}2 \text{, and thus }
\frac{g(u)}2\le\widehat g(u)\le\frac{3g(u)}2.
\end{equation}
We control the first term in~\eqref{eq:termstocontrol}:
\begin{align*}
\left|
\frac{\widehat a(x,u)}{\widehat g(u)}
-
\frac{a(x,u)}{g(u)}
\right|
&\le
\frac{|\widehat a(x,u)-a(x,u)|}{\widehat g(u)}
+
|a(x,u)|
\frac{|\widehat g(u)-g(u)|}{g(u)\widehat g(u)}\\
&\le
\frac{2\beta}{g(u)}
+
\frac{2|a(x,u)|\beta}{g(u)^2}\\
&\le
\frac{2\beta}{\alpha^K}
+
\frac{2|a(x,u)|\beta}{\alpha^{2K}}.
\end{align*}
For the second term in~\eqref{eq:termstocontrol}, we first control the denominator.
Put $M_K=B_P^{K+1}$, so that Lemma~\ref{lem:blockregularity} gives
\[
|f(x,u)|\le M_K \text{, }
|a(x,u)|\le M_K \text{,  and }
|b(u)|\le M_K.
\]
Since $\beta\le\alpha^K/2\le1$, we also have
\[
|\widehat b(u)|
\le |b(u)|+|\widehat b(u)-b(u)|
\le M_K+1.
\]
Therefore
\begin{align*}
|\widehat f(x,u)\widehat b(u)-f(x,u)b(u)|
&\qquad=
\left|
[\widehat f(x,u)-f(x,u)]\widehat b(u)
+
f(x,u)[\widehat b(u)-b(u)]
\right|\\
&\qquad\le
|\widehat f(x,u)-f(x,u)|\,|\widehat b(u)|
+
|f(x,u)|\,|\widehat b(u)-b(u)|\\
&\qquad\le
\beta(M_K+1)+M_K\beta
=
(2M_K+1)\beta.
\end{align*}
For the denominators,~\eqref{eq:termstocontrol2} implies
\begin{align*}
\left|
\frac1{\widehat g(u)^2}
-
\frac1{g(u)^2}
\right|
&=
\frac{
|\widehat g(u)-g(u)|[\widehat g(u)+g(u)]
}{
\widehat g(u)^2g(u)^2
}\\
&\le
\frac{\beta[5g(u)/2]}{[g(u)^2/4]g(u)^2}
=
\frac{10\beta}{g(u)^3}
\le
\frac{10\beta}{\alpha^{3K}}.
\end{align*}
Adding and subtracting $f(x,u)b(u)/\widehat g(u)^2$, we obtain the control of the second term in~\eqref{eq:termstocontrol}:
\begin{align*}
&\left|
\frac{\widehat f(x,u)\widehat b(u)}{\widehat g(u)^2}
-
\frac{f(x,u)b(u)}{g(u)^2}
\right|\\
&\qquad\le
\frac{
|\widehat f(x,u)\widehat b(u)-f(x,u)b(u)|
}{
\widehat g(u)^2
}
+
|f(x,u)b(u)|
\left|
\frac1{\widehat g(u)^2}-\frac1{g(u)^2}
\right|\\
&\qquad\le
\frac{4(2M_K+1)\beta}{\alpha^{2K}}
+
\frac{10M_K^2\beta}{\alpha^{3K}}.
\end{align*}
Combining the controls of both terms in~\eqref{eq:termstocontrol}, we obtain
\begin{align*}
\left|
\bigl[\widehat D_{j,K,h}(x,u)\bigr]_r
-
\partial_{j,r}p_K(x\mid u)
\right|
&\le
\left[
\frac{2}{\alpha^K}
+
\frac{10M_K+4}{\alpha^{2K}}
+
\frac{10M_K^2}{\alpha^{3K}}
\right]\beta\\
&\le
26M_K^2\alpha^{-3K}\beta.
\end{align*}
This bound holds for every $(x,u)\in I_{K,h}$ and every
$r\in\{1,\ldots,d\}$. Passing from scalar components to the
Euclidean norm therefore gives
\[
\sup_{(x,u)\in I_{K,h}}
\left\|
\widehat D_{j,K,h}(x,u)-\nabla_jp_K(x\mid u)
\right\|_2
\le
26\sqrt d\,M_K^2\alpha^{-3K}\beta.
\]
Finally, $M_K=B_P^{K+1}$ and $\beta=A_P^{K+1}t$.
All factors multiplying $t$ are bounded by an exponential in
$K+1$. Thus there is a constant $R_P\ge2$, independent of
$n,K,h,\eta$, such that
\[
\sup_{(x,u)\in I_{K,h}}
\left\|
\widehat D_{j,K,h}(x,u)-\nabla_jp_K(x\mid u)
\right\|_2
\le
R_P^{K+1}t
\qquad\text{for every }j\le K.
\]
Introduce the intermediate quantity
\[
L_{j,h}^{(K)}
=
\sup_{z\in I_{K,h}}\|\nabla_jp_K(z)\|_2.
\]
The reverse triangle inequality gives
\[
\left|\widehat L_{j,K,h}-L_{j,h}^{(K)}\right|
\le
\sup_{z\in I_{K,h}}
\|\widehat D_{j,K,h}(z)-\nabla_jp_K(z)\|_2
\le R_P^{K+1}t.
\]
This controls the statistical error on the interior. It remains
to compare $L_{j,h}^{(K)}$ with
\[
L_j^{(K)}
=
\sup_{z\in[0,1]^D}\|\nabla_jp_K(z)\|_2.
\]

For any $z\in[0,1]^D$, define $z^h$ by moving each scalar
coordinate to its nearest point in $[h,1-h]$:
\[
z^h_r=\min\{1-h,\max\{h,z_r\}\},
\qquad r=1,\ldots,D.
\]
Then $z^h\in I_{K,h}$ and
\[
|z_r-z^h_r|\le h,
\qquad
\sum_{r=1}^D|z_r-z^h_r|\le Dh.
\]
Every scalar component of $\nabla_jp_K$ has all its first
partial derivatives bounded by $B_P^{K+1}$, by the
second-derivative bound for $p_K$. Applying the mean value
inequality along the segment from $z$ to $z^h$ yields
\[
\|\nabla_jp_K(z)-\nabla_jp_K(z^h)\|_2
\le
\sqrt d\,B_P^{K+1}
\sum_{r=1}^D|z_r-z^h_r|
\le
\sqrt d\,D B_P^{K+1}h.
\]
Therefore
\[
\|\nabla_jp_K(z)\|_2
\le
L_{j,h}^{(K)}+\sqrt d\,D B_P^{K+1}h.
\]
Taking the supremum over the full cube, and observing that
the interior cube is a subset of it, gives
\[
0\le L_j^{(K)}-L_{j,h}^{(K)}
\le \sqrt d\,D B_P^{K+1}h.
\]

Combining the statistical and boundary errors, we obtain
\[
|\widehat L_{j,K,h}-L_j^{(K)}|
\le
R_P^{K+1}t+\sqrt d\,D B_P^{K+1}h.
\]
Since $h\le t$, summing gives
\[
\frac1\pi\sum_{j=1}^K
|\widehat L_{j,K,h}-L_j^{(K)}|
\le
\frac K\pi
\left[R_P^{K+1}+\sqrt d\,D B_P^{K+1}\right]t.
\]
Finally, $D=d(K+1)$, and the polynomial factors in $K$ can
be bounded by exponentials: for example,
$K\le2^K$ and $K(K+1)\le4^K$.
We may thus choose $C_P$ large enough that, for every $K\ge1$,
\[
\frac K\pi
\left[R_P^{K+1}+\sqrt d\,D B_P^{K+1}\right]
\le C_P^{K+1}.
\]
We conclude that
\[
\frac1\pi\sum_{j=1}^K
|\widehat L_{j,K,h}-L_j^{(K)}|
\le C_P^{K+1}t
=e_{n,K,h}(P,\eta),
\]
which proves the lemma.
\end{proof}

We are finally ready to prove the theorem.
\begin{proof}[Proof of Theorem~\ref{thm:continuousbound}]
If $e_{n,K,h}(P,\eta)\ge1$, the assertion is trivial since $\Delta_L(P)$ and its estimator borh belong to $[0,1]$. Thus, we consider only $e_{n,K,h}(P,\eta)<1$, so that we can apply Lemma~\ref{lem:ratio}. Together with Lemma~\ref{lem:feedback}, it yields, 
with probability at least $1-\eta$,
\[
 \abs{\frac1\pi\sum_{j=1}^K\widehat L_{j,K,h}-\frac L\pi}
 \le e_{n,K,h}(P,\eta)+\frac{B_K}\pi.
\]
Thus,
$$
|\widetilde\Delta^L_{n,K,h}-\dl(P)|
=
\abs{
{\rm clip}_{[0,1]}\left(1-\frac1\pi\sum_{j=1}^K\widehat L_{j,K,h}\right)
- \left(1-\frac L\pi\right)
}
\leq e_{n,K,h}(P,\eta)+\frac{B_K}\pi,
$$
which ends the proof.
\end{proof}

\begin{proof}[Proof of Corollary~\ref{cor:continuous}]
Write $\ell=\log(n+e)$. For all sufficiently large $n$, $K_n=\lfloor\ell^{1/3}\rfloor$ and $h_n=\exp\{-\ell/[4(D_n+3)]\}$. Fix $q>0$ and take $\eta=n^{-(q+2)}$. Since $D_n=d(K_n+1)$,
\begin{align*}
 C_P^{K_n+1}h_n
 &=\exp\left\{-\left(\frac1{4d}+o(1)\right)\ell^{2/3}\right\},\\
 C_P^{K_n+1}h_n^{-(D_n+2)}
       \sqrt{V_{n,K_n,h_n}(\eta)/n}
 &=n^{-1/4+o(1)}.
\end{align*}
The sum is eventually bounded by $\exp\{-c_d(\log n)^{2/3}\}$ with $c_d=1/[32(d+3)]$. The threshold depends on $P$ and $q$, but $c_d$ does not. Lemma~\ref{lem:feedback} shows that $KT_K^L\to0$ implies $B_K\to0$. Theorem~\ref{thm:continuousbound} therefore proves \eqref{eq:fixedaccuracy} for $\dl$. A summable failure level and Borel--Cantelli prove almost-sure convergence.
\end{proof}

\bibliographystyle{plainnat}

\end{document}